%% file: main.tex
\documentclass[]{logos_robotic_background}

\usepackage[toc,page,header]{appendix}

\usepackage{minitoc}
\usepackage{subcaption}
\usepackage{pifont}
\usepackage{xcolor,colortbl}
\usepackage{amsfonts}
\usepackage{caption}
\usepackage{float}
\usepackage{tabularx}
\usepackage{mathtools}

\usepackage{footmisc}
\usepackage{minted}
\usepackage{algpseudocode}
\usepackage{amssymb}
\usepackage{bm}
\usepackage{booktabs,siunitx,multirow, pifont}
\usepackage[dvipsnames]{xcolor}
\usepackage{siunitx}
\usepackage{tabularx}
\usepackage{makecell}
\usepackage{float}
\usepackage[table]{xcolor}
\usepackage{algorithm}
\usepackage{listings}
\usepackage{placeins} 
\usepackage[most]{tcolorbox}

\newcommand{\cmark}{\ding{51}}

\newcommand{\xmark}{\ding{55}}

\usepackage{listings}
\usepackage{xcolor}
\usepackage{caption}
\definecolor{codegreen}{rgb}{0,0.6,0}
\definecolor{codegray}{rgb}{0.5,0.5,0.5}
\definecolor{codepurple}{rgb}{0.58,0,0.82}
\definecolor{backcolour}{rgb}{0.95,0.95,0.92}
\lstdefinestyle{mystyle}{
    commentstyle=\color{codegreen},
    keywordstyle=\color{magenta},
    numberstyle=\tiny\color{codegray},
    stringstyle=\color{codepurple},
    basicstyle=\footnotesize\ttfamily,
    breakatwhitespace=false,         
    breaklines=true,                 
    captionpos=b,                    
    keepspaces=true,                 
    numbers=left,                    
    numbersep=5pt,                   
    showspaces=false,                
    showstringspaces=false,
    showtabs=false,                  
    tabsize=2,
    frame=none,                    
    framesep=3mm,
    rulecolor=\color{black},
    language=Python,
    morekeywords={I, f, f_new, hfov, wfov, W, H, W_new, H_new, I_new, resize, atan, tan}
}
\usepackage{makecell}

\title{Reinforcing Action Policies by \textit{Prophesying}}

\author[1\,3\,*]{Jiahui Zhang}
\author[1\,3\,*]{Ze Huang}
\author[1\,2\,3]{Chun Gu}
\author[1\,2]{Zipei Ma}
\author[1\,2\,3\,\S]{Li Zhang}

\affiliation[1]{School of Data Science, Fudan University}
\affiliation[2]{Shanghai Innovation Institute}
\affiliation[3]{Logos Robotics}

\contribution{Equal contribution.}

\newcommand{\wm}{\textit{Prophet}}
\newcommand{\rl}{\textit{FlowScale}}
\newcommand{\prl}{\textit{ProphRL}}
\newcommand{\fa}{\textit{FA-GRPO}}
\newcommand{\deltastep}[2]{\textcolor{ForestGreen}{\footnotesize\,(#1)}{\scriptsize~/#2}}
\newcommand{\deltaup}[1]{\textcolor{ForestGreen}{\footnotesize\,(#1)}}
\newcommand{\deltadown}[1]{\textcolor{BrickRed}{\footnotesize\,(#1)}}

\abstract{
Vision–Language–Action (VLA) policies excel in aligning language, perception, and robot control.
However, most VLAs are trained purely by imitation, which overfits to demonstrations, and is brittle under distribution shift. 
Reinforcement learning (RL) directly optimizes task reward and thus addresses this misalignment, but real-robot interaction is expensive and conventional simulators are hard to engineer and transfer. 
We address both data efficiency and optimization stability in VLA post-training via a learned world model and an RL procedure tailored to flow-based action heads.
Specifically, we first introduce \textbf{\wm{}}, a unified action-to-video robot world model pretrained on large-scale, heterogeneous robot data to learn reusable action–outcome dynamics and then few-shot adapted to new robots, objects, and environments, yielding a rollout-ready simulator.
Upon \wm{}, we reinforce action policies with our proposed \rl{}, which couples \textit{Flow-GRPO} with intrinsic stepwise reweighting to stabilize gradients.
Together, our solution provides a practical, data- and compute-efficient path to VLA post-training.
Experiments show 5--17\% success gains on public benchmarks and 24--30\% on real robots across diverse VLA backbones.
}

\checkdata[Correspondence $\S$]{Li Zhang at \email{lizhangfd@fudan.edu.cn}}

\checkdata[Project page]{\url{https://LogosRoboticsGroup.github.io/ProphRL}}

\begin{document}
{%
\renewcommand\twocolumn[1][]{#1}%
\maketitle
}

\input{sections/1-introduction}

\input{sections/2-relatedwork}
\input{sections/3-method}
\input{sections/4-experiments}
\input{sections/5-conclusion}

\bibliographystyle{plainnat}
\setlength{\bibsep}{0pt}
\bibliography{main}

\end{document}

%% file: sections/1-introduction.tex
\section{Introduction}
\label{sec:intro}

Vision--Language--Action (VLA) policies~\cite{black2025pi_, black2410pi0, zhang20254d, kim2025openvla, shukor2025smolvla, wang2025vla, qu2025spatialvla, zhang2025dreamvla} control robots from language instructions and visual observations, and now operate over image sequences with multistep action generation~\cite{wang2025vla,black2025pi_,kim2025openvla} rather than single-frame instruction following.
Despite progress, standard VLA training remains imitation-heavy and suffers from objective misalignment, i.e., likelihood-based objectives do not directly optimize long-horizon task reward.
Policies therefore become brittle under distribution shift and accumulate errors over long horizons.

To mitigate these issues, recent works add reinforcement learning (RL) post-training to VLA policies~\cite{sheng2025hybridflow,ouyang2022training,li2025simplevla,chen2025tgrpo,lu2025vla}, optimizing task-reward--aligned objectives rather than demonstration likelihood.
In practice, however, online RL in robotics faces high interaction cost, limited parallelism, and human-in-the-loop overhead.
Classical simulators require substantial engineering and often show domain transfer gaps for RGB-based policies.
Offline RL, in turn, lacks closed-loop data from the current policy, weakening long-horizon credit assignment.
Data-driven world models offer a middle ground: they generate action-conditioned futures at scale, enabling policies to practice in imagination while reusing the same visual interface as VLAs.

However, most existing efforts~\cite{liao2025genie, guo2025ctrl, jang2025dreamgen, zhu2024irasim, jiang2025enerverse, liang2025video, team2025gigabrain} remain confined to single-scene world models and, even when paired with VLA policies, primarily use them as data augmentors rather than truly adaptable simulators, leaving open how such models generalize to new real-world scenes and goals. 
A few works~\cite{xiao2025world,li2025vla,2025wmpo,jiang2025world4rl} that do employ world models as simulators for VLA RL largely focus on replacing an existing simulator with a learned one, and stop short of addressing the harder question of using world models as practical RL backends for real-world robotics.
This leaves open the core question of how to acquire a general-purpose, few-shot adaptable world model that remains useful in the real world. 

In this work, we argue that a more practical goal is to learn a world model that can adapt with few shots to new embodiments, tasks, and scenes under realistic data and compute budgets.
Such a rollout-ready model provides a practical substrate for RL, delivering action-aligned long-horizon feedback without physical risk.
To this end, we present \wm{}, an action-conditioned world model built upon a video generator that predicts long-horizon robot manipulation rollouts from first-frame observation and multi-step action inputs.
Our \wm{} is pretrained over diverse robot manipulation datasets to learn general action-to-outcome dynamics, then adapted with few shots to new situations.
To assess action-conditioning fidelity, we also introduce an optical flow-guided evaluation protocol that measures end-effector accuracy and interaction fidelity beyond conventional video quality metrics.

Within this substrate, we develop \prl{}, which couples \wm{} with \fa{} and \rl{} to reinforce VLA policies:
\fa{} aligns Flow-GRPO~\cite{liu2025flow} ratios with environment-level actions rather than individual flow steps, 
while \rl{} reweights per-step gradients using the noise schedule to reduce score-driven heteroscedasticity and stabilize updates.

\input{images/teaser}

Our contributions are as follows:
\textbf{(i)} We introduce \wm{}, a history-aware, action-conditioned world model that produces action-aligned, long-horizon manipulation rollouts and interfaces directly with the VLA observation space;
\textbf{(ii)} We conduct large-scale pretraining across heterogeneous robot datasets followed by few-shot adaptation to new environments, objects, and trajectories, yielding a rollout-ready world simulator;
\textbf{(iii)} We propose an optical flow-guided evaluation protocol that assesses end-effector trajectories and interaction fidelity, complementing conventional quality metrics;
\textbf{(iv)} We develop \fa{} and \rl{}, an RL post-training scheme tailored to flow-based action heads for VLA policies.

%% file: images/teaser.tex
\begin{figure}
    \centering
    \includegraphics[width=1.0\linewidth]{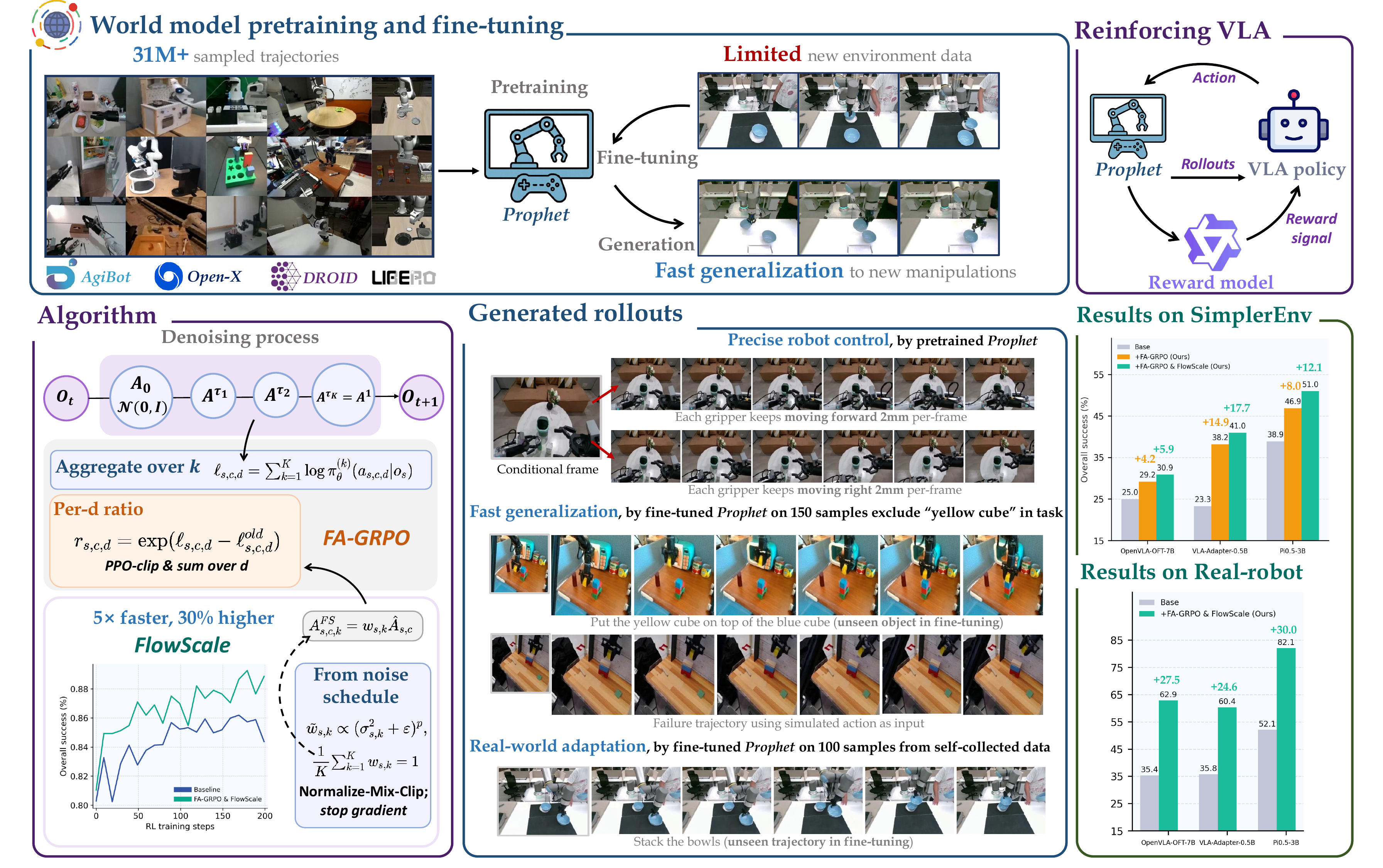}
\caption{\textbf{We propose a framework that uses a world model as a real-world simulator to post-train VLA policies.}
    Our world model \wm{} is pretrained on large-scale robot trajectories to model action-to-video dynamics.
    The pretrained \wm{} enables `prophesying' precise, physically plausible long-horizon rollouts, and can be rapidly adapted via few-shot fine-tuning to new environments, objects, and trajectories.
    Upon \wm{}, we introduce the \fa{} with \rl{} RL algorithm to more stably and efficiently improve policies.
    Together, our training paradigm turns diverse logged data and a single pretrained world model into a unified engine for scalable, data-efficient, and safely improvable VLA systems.}
    \label{fig:teaser}
\end{figure}

%% file: sections/2-relatedwork.tex
\section{Related Works}

\noindent \textbf{World models for robot manipulation}
Prior work uses text-conditioned video generation~\cite{du2023learning, bruce2024genie, quevedo2025worldgym}, sometimes with coarse action cues such as move left or go up~\cite{quevedo2025worldgym, yang2023learning}. 
These models leverage web-scale priors for language grounding and scene understanding, but their loose conditioning limits controllability, making them ill suited for manipulation policies that require precise geometry and reliable dynamics. 
More recent works~\cite{liao2025genie, guo2025ctrl, jang2025dreamgen, zhu2024irasim, jiang2025enerverse, liang2025video, team2025gigabrain} condition on low-level robot signals, such as end-effector poses or joint trajectories, to generate future manipulation videos and sometimes serve as simulators for downstream control~\cite{li2025vla, xiao2025world, jiang2025world4rl, 2025wmpo}. 
Yet they still struggle with long-horizon rollouts and realistic failure modes, and often require substantial new data to adapt beyond training domains. 
Evaluation typically relies on generic video metrics rather than checking action-conditioned correctness of end-effector motion and contact, leaving their utility for control-centric applications unclear.

\noindent \textbf{RL for VLA policies}
Policy-gradient methods~\cite{williams1992simple,schulman2017proximal,schulman2015trust,lillicrap2015continuous,haarnoja2018soft,fujimoto2018addressing} have long guided on-policy robotics under data and safety constraints~\cite{qi2025sofar, zhang2022predict}. 
Across foundation models, RL algorithms~\cite{shao2024deepseekmath,yu2025dapo,hu2025reinforce++,cui2025process,li2023remax,liu2025flow} enabled large-scale post-training on outcome-aligned signals. 
Recent works~\cite{li2025simplevla,lu2025vla} transfer these recipes to RL post-training, typically optimizing log-likelihood objectives on image-conditioned action sequences. 
For VLA policies~\cite{black2410pi0,intelligence2504pi0,shukor2025smolvla,cheang2025gr,lv2025f1} with flow-based action heads, several studies~\cite{zhang2025reinflow,li2025vla} modify sampling to make flows RL compatible but leave the update signal unchanged, so score-driven gradient heteroscedasticity persists and long-horizon stability remains limited, often alongside reliance on task-specific simulators. 

%% file: sections/3-method.tex
\section{Methodology}

\subsection{Overall training paradigm}
\begin{figure}
    \centering
    \includegraphics[width=0.73\linewidth]{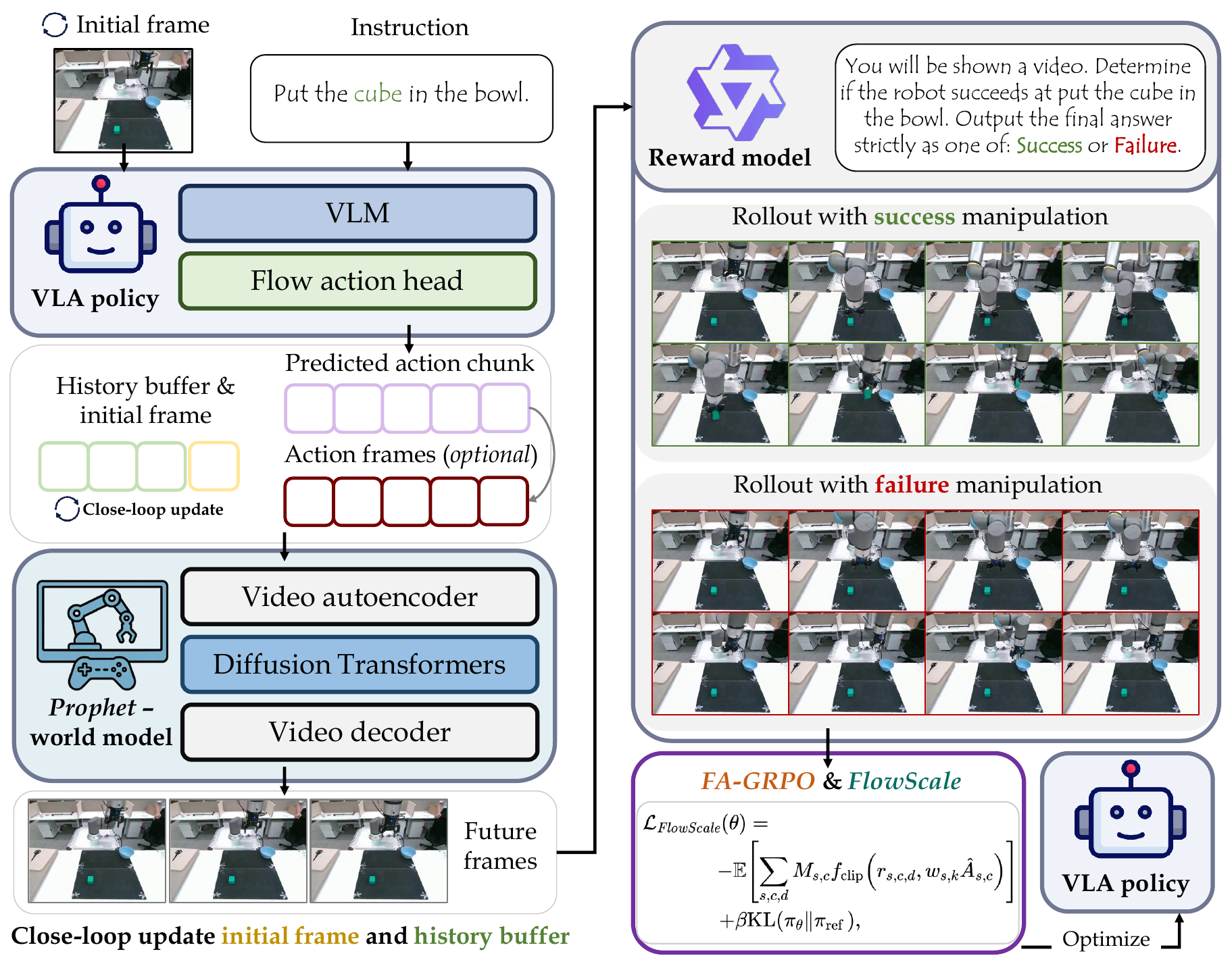}
\caption{\textbf{\prl{} Training paradigm.} Given an initial frame and instruction, the policy predicts an action chunk and \wm{} generates the future robot rollout, updating the policy input, current frame, and history buffer until the episode ends. An offline reward model scores each full trajectory, and the policy is reinforced with \fa{} and \rl{} using these `prophesied' and realistic rollouts.}
    \label{fig:pipe}
\end{figure}

Our overall training paradigm, \prl{}, couples \wm{} with VLA post-training, as shown in Fig.~\ref{fig:pipe}.
At each outer step, the policy predicts an action chunk from instruction and initial image.
\wm{} generates a clip conditioned on this chunk and the initial image. The clip is fed back to the policy and \wm{}, enabling long-horizon closed-loop rollouts.
A frozen VLM-based reward model~\cite{bai2025qwen2} scores each rollout to produce group rewards, using a prompt template.
Finally, we optimize the policy with \fa\ (Eq.~\eqref{eq:fa}) using group-normalized advantages, while \rl\ (Eq.~\eqref{eq:flowscale}) reweights denoising steps to stabilize gradients.

\subsection{World model}
\noindent \textbf{Preliminaries}
We build \wm{} on a latent video diffusion pipeline.
A video autoencoder encodes a real clip $x_{1:T}$ into latents $z_0=\mathrm{E}(x_{1:T})$ and approximately reconstructs it via $\hat x_{1:T}\approx\mathrm{D}(\mathbf{z}_0)$.
A DiT denoiser then learns a conditional diffusion model over $\mathbf{z}_0$: at step $t$ we add Gaussian noise to obtain $\mathbf{z}_t$, and the denoiser takes $(\mathbf{z}_t, t, f)$, where $f$ denotes conditioning, to iteratively predict a clean latent $\hat{\mathbf{z}}_0$ that is finally decoded back to video.
The denoiser is optimized with the standard latent-space noise prediction objective:
\begin{equation}
\mathcal{L}_{\text{diff}}=\mathbb{E}_{\mathbf{z}_0,\epsilon\sim\mathcal{N}(0,\mathbf{I}),t} \Big[\left \| \epsilon - \epsilon_\theta\big(\mathbf{z}_t,t,f\big) \right \|_{2}^{2} \Big],
\end{equation}
where $\mathbf{z}_t=\sqrt{\bar{\alpha}_t}\mathbf{z}_0+ \epsilon\sqrt{1-\bar{\alpha}_t}$.
$\mathbf{z}_0$ is the clean latent from the encoder, $t$ is a diffusion step, $f$ is the conditioning feature, and $\epsilon_\theta$ predicts the injected noise, teaching the model to invert the noising process under conditioning.

\subsubsection{Definition of action}
\label{app:definition_action}
For each trajectory we represent the low-level control commands as $c \in \mathbb{R}^{T \times N \times D}$, where $T$ is the time horizon, $N$ is the number of end-effectors, and $D$ is the action dimension.
To enable cross-dataset pretraining, we fix $N$ to be the maximum number of end-effectors across all datasets (i.e., 2 on AgiBot~\cite{bu2025AgiBot}), and pad trajectories with fewer end-effectors with zeros along the end-effector dimension. 
These padded entries do not correspond to any physical end-effector but allow us to keep a single, shared action parameterization.

Each per-step, per-end-effector action is a 7-dim vector (i.e., $D=7$):
\begin{equation}
    c_{t,n} = \big[\Delta p_{t,n}^\top,\Delta e_{t,n}^\top,g_{t,n}\big]^\top\in \mathbb{R}^7, \quad t = 1,\dots,T, \quad n = 1,2,
\end{equation}
where $\Delta p_{t,n} \in \mathbb{R}^3$ is a translational delta, $\Delta e_{t,n} \in \mathbb{R}^3$ is a rotational delta expressed in Euler angles, and $g_{t,n} \in [0, 1]$ denotes the open ratio of the gripper.

During large-scale pretraining, we parameterize the motion as a local delta pose with respect to the previous end-effector frame. 
Let $\xi_{t-1,n} \in \mathrm{SE}(3)$ denote the transform of end-effector $n$ at time $t-1$ in the world coordinate system. 
The action $c_{t,n}$ corresponds to a target transform $\xi_{t,n}$ obtained by applying a small rigid-body motion $\Delta \xi_{t,n}$ in the local frame of $\xi_{t-1,n}$:
\begin{equation}
    \Delta \xi_{t,n} =\big(\Delta p_{t,n}, \Delta e_{t,n}\big), \quad
\xi_{t,n} = \xi_{t-1,n} \circ \Delta \xi_{t,n},
\end{equation}
where $\Delta p_{t,n}$ and $\Delta e_{t,n}$ are interpreted as a
translation and a rotation around the end-effector origin at time
$t-1$. 
This delta formulation makes the action space more homogeneous across tasks and datasets.

For fine-tuning, we retain the same 7-dim structure but adapt the
semantics of the deltas to match the low-level controller used in each environment. 
For example, in simulator-based LIBERO, the action is interpreted as a servo command directly consumed by the environment, whereas on real-robot and BRIDGE the action corresponds to the difference between consecutive absolute poses (i.e., $\xi_{t,n} - \xi_{t-1,n}$ in position and Euler angle). 
In all cases, the gripper $g_{t,n}$ remains a normalized scalar indicating the desired degree of opening.

\begin{figure}
    \centering
    \includegraphics[width=0.85\linewidth]{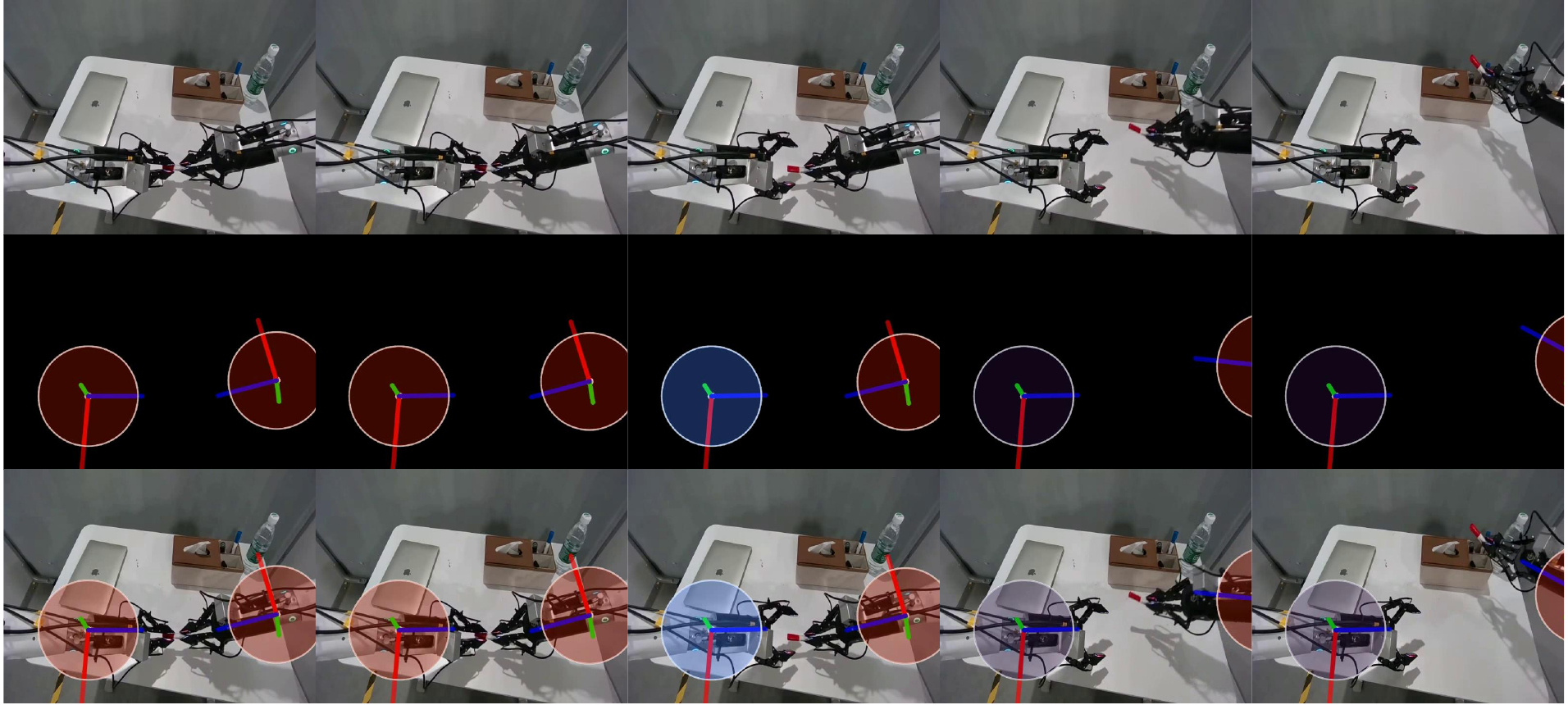}
    \caption{\textbf{Action frame visualization.} The first row shows RGB frames, the middle row shows the constructed action frames, and the last row shows the alignment between the visualized action frames and the image pixels.}
    \label{fig:action-frame}
\end{figure}

\subsubsection{Construct action frames}
To provide \wm{} with a compact yet geometry–aware representation of the robot motion, we construct an action frame by projecting the end-effector action onto the camera image plane and rendering a 2D visualization on a black background, following~\cite{liao2025genie, jiang2025enerverse}. 
For each time step $t$, we assume access to:
\textbf{(i)} the camera intrinsics $K \in \mathbb{R}^{3 \times 3}$;
\textbf{(ii)} the camera extrinsics $E_t \in \mathrm{SE}(3)$ that transform 3D points from the world to the camera coordinates;
\textbf{(iii)} the end-effector position $p_{t,j} \in \mathbb{R}^3$ and rotation $\mathbf{R}_{t,j} \in \mathbb{R}^{3 \times 3}$ for each end-effector $j$;
\textbf{(iv)} a scalar control signal $g_{t,j}$ (i.e., gripper opening degree) used to encode the action magnitude.

Let $\{n^x, n^y, n^z\}$ denote three axis-aligned unit vectors in the end-effector frame, scaled by a constant axis length $l$:
\begin{equation}
    n^k = l\,v^k,\quad k \in \{x,y,z\},
\end{equation}
where $v^k$ is the $k$-th canonical basis vector. 
The corresponding 3D points in the world frame consist of the end-effector origin and the tips of the three axes 
\begin{equation}
    p^0_{t,j} = p_{t,j},\quad p^k_{t,j} = p_{t,j} + \mathbf{R}_{t,j} n^k,\quad k \in \{x,y,z\}.
\end{equation}
These points are transformed into the camera coordinates:
\begin{equation}
    x^k_{t,j} = E_t
    \begin{bmatrix}
    p^k_{t,j} \\ 1
    \end{bmatrix}
    \in \mathbb{R}^3,
\end{equation}
and then projected onto the image plane:
\begin{equation}
    \tilde{u}^k_{t,j} = K\,x^k_{t,j},\quad
    u^k_{t,j} = 
    \left(
    \frac{\tilde{u}^k_{t,j,x}}{\tilde{u}^k_{t,j,z}},
    \frac{\tilde{u}^k_{t,j,y}}{\tilde{u}^k_{t,j,z}}
\right),
\end{equation}
where $u^k_{t,j}$ are 2D pixel coordinates and $z^0_{t,j} = x^0_{t,j,z}$ denotes the depth of the end-effector origin in the camera frame.
To make the visualization depth-aware, we set the radius of the rendered disk to be a monotonically decreasing function of the depth:
\begin{equation}
\mathbf{r}(z^0_{t,j}) = \operatorname{clip}\!\left(
\mathbf{r}_{\text{ref}} \cdot \frac{z_{\text{ref}}}{z^0_{t,j}}, \, \mathbf{r}_{\min},\, \mathbf{r}_{\max}
\right),
\end{equation}
where $\mathbf{r}_{\text{ref}}$ is the radius at a reference depth $z_{\text{ref}}$, and $\mathbf{r}_{\min}, \mathbf{r}_{\max}$ bound the radius range. 
We also map the scalar control signal $g_{t,j}$ to a color using a fixed colormap $\text{CM}(\cdot)$ defined over a value range $[g_{\min}, g_{\max}]$:
\begin{equation}
   \text{color}_{t,j} = \text{CM}(g_{t,j}) \in \mathbb{R}^3. 
\end{equation}
Finally, we render the action frame on a black canvas $\mathrm{c}_t \in \mathbb{R}^{H \times W \times 3}$. 
For each end-effector $j$, we draw a filled disk with radius $r(z^0_{t,j})$ and color $\text{color}_{t,j}$ centered at $u^0_{t,j}$, and overlay a small white point at the center and three colored line segments from $u^0_{t,j}$ to $u^k_{t,j}$, $k\in\{x,y,z\}$, indicating the local orientation. 

In practice, we draw the colored disk onto a separate overlay and alpha-blend it with the line drawing to obtain a smooth appearance.
Fig.~\ref{fig:action-frame} visualizes the action frames constructed on AgiBot~\cite{bu2025AgiBot}.

\subsubsection{Action conditioning}
Our \wm{} conditions on actions at two levels: 
\textbf{(i)} a global chunk-level embedding of the scalar action stream, and 
\textbf{(ii)} an optional latent embedding of the action frames.

\noindent\textbf{Notation}
In practice, each per-step action is a tensor of shape $(N, D)$ over $N$ end-effectors (Sec.~\ref{app:definition_action}), and we pad $N$ to the maximum number of end-effectors across datasets.
For notational simplicity, we fold the end-effector dimension and write actions as $c_{1:T} \in \mathbb{R}^{T \times D}$.
The explicit end-effector index is omitted to keep the action-conditioning formulas concise.

\noindent\textbf{Scalar action stream}
Given a per-step action sequence
$c_{1:T}\in\mathbb{R}^{T\times D}$, we first flatten the whole chunk into a single vector of shape $[T \times D]$ and map it to a global embedding via an MLP, $f_{\text{sa}} = \phi(c_{1:T}) \in \mathbb{R}^{D_{m}}$,
where $D_{m}$ is the DiT channel dimension.

Let $\tilde{t} \in \mathbb{R}^{T_{l} \times D_{m}}$ denote the standard timestep embeddings produced by the DiT time-embedder.
We inject the scalar action conditioning by simply adding the global action embeddings, $\tilde{t} = \tilde{t} + f_{\text{sa}}$.
Broadcasting over the temporal dimension is applied in practice.

\noindent\textbf{Action frame stream}
When action frames $\mathrm{c}_{1:T}\in\mathbb{R}^{T\times H\times W\times 3}$ are available, we additionally condition on their latent representation. 

Let $\mathbf{z}_{\text{af}}=\mathrm{E}(\mathrm{c}_{1:T}) \in \mathbb{R}^{T_{l}\times C_{l} \times H_{l} \times W_{l}}$ denote the latent action video obtained by encoding action frames.
The lightweight 3D projection $\psi$ first maps $\mathbf{z}_{\text{af}}$ into the DiT channel space $D_{m}$ via a $1\times1\times1$ convolution, followed by depthwise separable $1\times3\times3$ and $1\times1\times1$ convolutions over space:
\begin{equation}
    \mathbf{u} = \mathrm{Conv}_{1\times1\times1}(\mathbf{z}_{\text{af}}), \quad 
    \mathbf{u} = \mathrm{DWConv}_{1\times3\times3}(\mathbf{u}), \quad
    \mathbf{u} = \mathrm{PWConv}_{1\times1\times1}(\mathbf{u}).
\end{equation}
We then average-pool $\mathbf{u} \in \mathbb{R}^{T_{l}\times D_{m} \times H_{l} \times W_{l}}$ over the spatial dimensions and add a sinusoidal positional encoding along temporal dimension:
\begin{equation}
    \mathbf{h}_t = \mathrm{AvgPool}_{H_{l},W_{l}}(\mathbf{u}_t) + \mathrm{PE}(t),
\quad \mathbf{h}_t \in \mathbb{R}^{D_{m}}.
\end{equation}
Finally, an MLP maps $\mathbf{h} \in \mathbb{R}^{T \times D_{m}}$ to the final action frame conditioning feature $f_{\text{af}} \in  \mathbb{R}^{T_{l} \times D_{m}}$, which is added to the timestep embeddings $\tilde{t} = \tilde{t} + f_{\text{sa}} + f_{\text{af}}$.

\subsubsection{History-aware mechanism}
We maintain a low-resolution memory over past latent frames using a FramePack-style~\cite{zhang2025framepack, zhang2025framepackv1} module. 

Given a history latent $\mathbf{z}^{\text{hist}} \in \mathbb{R}^{T_{\mathrm{h}} \times C_l \times H_{l} \times W_{l}}$ computed by video autoencoder from the history buffer $h_{-T_{h}:0}$, the history-aware module applies several 3D average-pooling and $1\times1\times1$ projection blocks at different spatio-temporal strides, and concatenates all resulting tokens into a memory matrix $M \in \mathbb{R}^{L_{\mathrm{h}} \times D_{m}}$.
Two linear layers map $M$ to additional key and value vectors $(\mathrm{K}_{\mathrm{mem}}, \mathrm{V}_{\mathrm{mem}})$, which are fed into all DiT blocks as an external concatenated memory $\tilde{Q}=\mathrm{Attn}(Q,[\mathrm{K}_{\text{mem}};\mathrm{K}],[\mathrm{V}_{\text{mem}};\mathrm{V}])$.

The history-aware mechanism provides long-range temporal context for stable geometry and contact evolution while keeping computation predictable.

\subsubsection{Long rollout generation}
We generate long videos autoregressively in chunks.
Starting from the first observed frame, we initialize the history buffer with this frame. 
Given an action rollout for the first segment, the model produces a short clip. 
The last generated frame is then used as the start frame for the next segment, and the newly generated clip is compressed into the history. 
We repeat this procedure to maintain temporal continuity while scaling to long horizons.
A detailed long rollout generation with history buffer updating procedure is shown in Alg.~\ref{alg:long_rollout}.

\begin{algorithm}[t]
\caption{Closed-loop rollout with \wm{} given streaming action chunks}
\label{alg:long_rollout}
\begin{algorithmic}[1]
\Require History length $T_h$, initial RGB frame $x_0$, chunk size $C$
\State Initialize history buffer $H \gets [x_0, x_0, \dots, x_0]$ (length $T_h$)
\For{$s = 1, 2, 3, \dots$} \Comment{iterate over incoming action chunks}
    \State Receive current action chunk $\mathcal{A}_s = [c^{(s)}_1,\dots,c^{(s)}_C]$
    \State $\mathcal{H} \gets$ last $T_h$ frames in $H$ \Comment{history input}
    \State $\hat{x}_{1:C} \gets \textit{Prophet}(x_0, \mathcal{A}_s, \mathcal{H})$
           \Comment{predict $C$ future frames}
    \State $x_0 \gets \hat{x}_C$ \Comment{last generated frame as next start frame}
    \State Append $[\hat{x}_1,\dots,\hat{x}_C]$ to $H$
    \State Keep only the most recent $T_h$ frames in $H$
    \If{no more action chunks are provided}
        \State \textbf{break}
    \EndIf
\EndFor
\end{algorithmic}
\end{algorithm}

\subsubsection{Optical flow-guided evaluation protocol}
Prior action-conditioned world models are usually evaluated only with video metrics (e.g., PSNR), which capture perceptual fidelity but not whether actions are executed correctly or physical interactions follow the intended control.
To address this, we introduce an optical flow–based protocol that compares motion fields between real videos and action-conditioned rollouts.
Given a real video $x_{1:T}$ and a rollout $\hat{x}_{1:T}$, we compute dense optical flow between consecutive frames using the Farnebäck estimator~\cite{farneback2003two} after grayscale conversion $\mathbf{u}_t=\mathrm{Flow}(x_t,x_{t+1}), \quad \hat{\mathbf{u}}_t=\mathrm{Flow}(\hat{x}_t, \hat{x}_{t+1})$.
We measure magnitude agreement with the end-point error:
\begin{equation}
    \mathrm{EPE}_t=\frac{1}{HW}\sum_x \lVert \mathbf{u}_t(x)-\hat{\mathbf{u}}_t(x)\rVert_2,
\end{equation}
and directional alignment with the cosine similarity:
\begin{equation}
    \cos_t=\frac{1}{|V_t|}\sum_{x\in V_t} \frac{\langle \mathbf{u}_t(x),\hat{\mathbf{u}}_t(x)\rangle}{\lVert\mathbf{u}_t(x)\rVert_2 \lVert \hat{\mathbf{u}}_t(x)\rVert_2+\varepsilon},
\end{equation}
where $V_t=\left \{ m:\lVert \mathbf{u}_t(x)\rVert_2>\tau \ \wedge\ \lVert \hat{\mathbf{u}}_t(x)\rVert_2>\tau \right \}$ filters near-static pixels, $\tau$ is a small threshold.
We report the mean and median endpoint error $\overline{\mathrm{EPE}}$, $\widetilde{\mathrm{EPE}}$, the mean and median flow-direction cosine $\overline{\cos}$, $\widetilde{\cos}$, aggregated over $t$, which jointly capture control-relevant motion magnitude and direction.
Such metrics compare motion fields rather than appearance, yielding an appearance-invariant, control-relevant assessment of whether the conditioned actions induce the correct end-effector and contact dynamics.

\subsection{RL-based VLA post-training}
\label{sec:rl_setup}

\noindent\textbf{Trajectory layout}
We represent a batch of episodic rollouts as a tensor of shape $[B,S,K,CH,D]$, where $B$ is the batch size, $S$ indexes outer model inference steps (environment steps and policy calls along the trajectory), $K$ is the number of denoising steps in the flow-based action head, $CH$ is the number of action chunks emitted per policy call, and $D$ is the action dimension per chunk.
For each episode $i\in\{1,\dots,B\}$ and outer step $s\in\{1,\dots,S\}$, the policy receives an observation $o^{(i)}_s$ and outputs a chunked action $\{a^{(i)}_{s,c,d}\}_{c=1,\dots,CH;\,d=1,\dots,D}$.
Each pair $(s,c)$ corresponds to one low-level control command executed sequentially between two policy calls, in all our experiments $D=7$, encoding end-effector translation, rotation, and a scalar gripper command.
The chunk index $c$ lets the policy emit a short open-loop sequence of commands per observation $o^{(i)}_s$, which empirically stabilizes long-horizon control.

\noindent\textbf{Flow action head and internal steps}
The flow-based action head factorizes the per-chunk, per-dimension log-likelihood across internal steps:
\begin{equation}
\log \pi_\theta(a_{s,c,d}\!\mid o_s)
= \sum_{k=1}^{K}\log \pi^{(k)}_\theta(a_{s,c,d}\!\mid o_s),
\label{eq:factor}
\end{equation}
where $\pi^{(k)}_\theta$ is the likelihood factor contributed by internal step $k$.
The index $k$ is internal to the flow head and does not advance environment time: for fixed $(s,c)$, the $K$ denoising updates all condition on $o_s$ and jointly parameterize a single environment-level action $a_{s,c}$.
In our RL objective, we first aggregate over $k$ as in~\eqref{eq:factor}, keep dimensions $d$ factorized, and treat each pair $(s,c)$ as one environment action. PPO-style ratios are computed per triplet $(s,c,d)$.

\noindent\textbf{Variable-length episodes and masking}
Episodes have variable lengths.
Let $T_i$ be the horizon of episode $i$ in outer steps.
We store trajectories in tensors of fixed length $S$ and pad remaining slots with dummy transitions.
Early termination is handled by a binary mask:
\begin{equation}
M_{s,c}^{(i)} \in \{0,1\}, \qquad
M_{s,c}^{(i)} =
\begin{cases}
1, & s \le T_i,\\
0, & s > T_i,
\end{cases}
\end{equation}
which zeroes out all loss contributions after the episode ends.
In practice, we broadcast $M_{s,c}^{(i)}$ over $k$ and $d$, and multiply both policy losses and advantages by this mask.
If an episode terminates between chunks, we conservatively set $M_{s,c}^{(i)} = 0$ for all chunks $c$ after the first terminal chunk, so that no gradient is propagated beyond the first invalid action.

\subsubsection{Flow-action-GRPO (\fa{})}
Vanilla Flow-GRPO~\cite{liu2025flow} treats each internal flow step $k$ as an atomic action and constructs PPO ratios per $(s,c,k)$ before summing over $k$.
To better match the environment, \fa{} instead aggregates all internal flow steps into an action-level log-probability and then forms ratios per dimension of each action chunk, i.e., at the level of $(s,c,d)$ while still treating each $(s,c)$ as one environment action.

Recall from Eq.~\eqref{eq:factor} that the flow-based action head factorizes the per-chunk, per-dimension log-likelihood across internal steps.
For each environment step $s$, chunk $c$, and action dimension $D$, we define the action-level log-probabilities under the current and behavior policies:
\begin{equation}
\begin{aligned}
\ell_{s,c,d}
&= \log \pi_\theta(a_{s,c,d}\!\mid o_s)
 = \sum_{k=1}^{K}\log \pi^{(k)}_\theta(a_{s,c,d}\!\mid o_s),\\
\ell^{\mathrm{old}}_{s,c,d}
&= \log \pi_{\mathrm{old}}(a_{s,c,d}\!\mid o_s),
\end{aligned}
\end{equation}
and a per-dimension PPO ratio:
\begin{equation}
r_{s,c,d}
= \exp\!\big(\ell_{s,c,d}-\ell^{\mathrm{old}}_{s,c,d}\big)
= \frac{\pi_\theta(a_{s,c,d}\!\mid o_s)}{\pi_{\mathrm{old}}(a_{s,c,d}\!\mid o_s)}\,.
\label{eq:ell-scd}
\end{equation}

Given action-level advantages $\hat A_{s,c}$ (one advantage per outer step and chunk, broadcast over $d$), we optimize a clipped-ratio objective with a KL regularizer:
\begin{equation}
\begin{split}
\mathcal{L}_{\fa}(\theta)
&= -\,\mathbb{E}\!\Big[\sum_{s,c,d} M_{s,c}\,
f_{\text{clip}}\!\big(r_{s,c,d},\,\hat A_{s,c}\big)\Big]
 + \beta\,\mathrm{KL}\!\big(\pi_\theta\,\|\,\pi_{\text{ref}}\big),
\end{split}
\label{eq:fa}
\end{equation}
where $f_{\text{clip}}(r,A)= \min\!\big\{rA,\, \mathrm{clip}(r,1-\varepsilon_\mathrm{low},1+\varepsilon_\mathrm{high})\,A\big\}$,
$\pi_{\text{ref}}$ is the frozen supervised VLA policy, and the mask $M_{s,c}$ zeroes out contributions from padded or terminated timesteps.
The KL term is evaluated on the same factorized per-dimension action distribution, i.e., over all $(s,c,d)$, and aggregated with the mask $M_{s,c}$.

Compared to the Flow-GRPO objective, Eq.~\eqref{eq:fa} only changes how internal flow steps are handled: instead of treating each $(s,c,k)$ as a separate action with its own ratio and advantage, we sum the log-likelihood contributions over $k$ into $\ell_{s,c,d}$ and use a single advantage $\hat A_{s,c}$ shared across all $d$ and $k$ for the same $(s,c)$.
This leaves the underlying stochastic policy over actions unchanged; it only changes how gradients from internal flow steps are aggregated, by broadcasting one scalar advantage per environment action over all dimensions and internal steps.

\subsubsection{Intrinsic stepwise reweighting (\rl)}
\label{sec:flowscale}

SDE-based flow heads exhibit highly non-uniform gradient magnitudes across internal steps $k$: early noisy steps and late refinement steps affect the overall action log-likelihood in very different ways.
Without any correction, low-noise steps (large $k$, small $t$) tend to dominate the update.
\rl{} introduces a state- and step-dependent weight $w_{s,k}$ that modulates each flow step’s contribution while keeping the action-centric surrogate of Eq.~\eqref{eq:fa} unchanged.

At the level of the scalar objective, we formulate the \rl{} loss as:
\begin{equation}
\begin{split}
\mathcal{L}_{\rl}(\theta)
&= -\,\mathbb{E}\!\Big[\sum_{s,c,d}  M_{s,c}\,
  f_{\text{clip}}\!\big(r_{s,c,d},\,w_{s,k}\hat A_{s,c}\big)\Big]
 + \beta\,\mathrm{KL}\!\big(\pi_\theta\,\|\,\pi_{\text{ref}}\big),
\end{split}
\label{eq:flowscale}
\end{equation}
where $r_{s,c,d}$ is the per-dimension PPO ratio from Eq.~\eqref{eq:ell-scd}.
Here $w_{s,k}$ should be understood as a stepwise modulation of the contribution of each internal flow step $k$ to the gradient of the aggregated log-probability, we treat it as a stop-gradient coefficient and do not change the underlying stochastic policy over actions.

\noindent\textbf{Per-step noise scale}
For each outer step $s$ and internal flow step $k$, we obtain a scalar noise scale from the diffusion/flow time schedule rather than predicting it with the network.
We use a noise schedule $\{\sigma_j\}_{j=0}^{K}$ and a normalized time variable $t_{s,k}\in[0,1]$.
The standard deviation of the injected noise is, up to a constant:
\begin{equation}
\texttt{std}_{s,k}
\;\propto\;
\sqrt{\sigma(t_{s,k})}\,\sqrt{|\Delta t|},
\label{eq:flowscale-std}
\end{equation}
implemented by combining a lookup table of $\sigma$ values with the current time $t_{s,k}$.
After squeezing channel dimensions we set:
\begin{equation}
\sigma_{s,k}^2 \coloneqq \texttt{std}_{s,k}^2
\quad\Rightarrow\quad
\sigma^2 \in \mathbb{R}^{B\times S\times K},
\end{equation}
and use $\sigma_{s,k}^2$ as a scalar proxy for the local noise level (or uncertainty) of the flow head at step $k$.

\begin{algorithm}[t]
\caption{\rl{} weight computation and application (per mini-batch)}
\label{alg:flowscale}
\begin{algorithmic}[1]
\State \textbf{Input:} log-probs $\texttt{logp\_elem}\in\mathbb{R}^{B\times S\times K\times CH\times D}$,
old log-probs $\texttt{old\_logp\_elem}$,
advantages $\hat A_{s,c}$,
action mask $M_{s,c}$,
per-step std $\texttt{std}\in\mathbb{R}^{B\times S\times K\times 1\times 1}$
\State \textbf{Hyperparameters:} $p$, $\alpha$, $w_{\min}$, $w_{\max}$, $\varepsilon$
\State $\sigma^2 \leftarrow \texttt{std.squeeze}(-1,-2)^2$ \hfill\Comment{$\sigma^2\in\mathbb{R}^{B\times S\times K}$}
\State $\tilde w \leftarrow (\sigma^2 + \varepsilon)^{p}$ \hfill\Comment{power-scaled weights}
\State $\bar w \leftarrow \tilde w / \text{mean}_k(\tilde w)$ \hfill\Comment{normalize so $\frac{1}{K}\sum_k \bar w_{s,k}=1$}
\State $w^{\text{mix}} \leftarrow \alpha \cdot 1 + (1-\alpha)\,\bar w$
\State $w \leftarrow \mathrm{clip}(w^{\text{mix}}, w_{\min}, w_{\max})$
\State $w \leftarrow \texttt{stop\_gradient}(w)$
\State Broadcast $w$ to shape $[B,S,K,CH,D]$ and multiply into $\hat A_{s,c}$ (or equivalently into $\log\pi^{(k)}_\theta$)
\State Compute the \fa{} loss of Eq.~\eqref{eq:fa} using the weighted advantages and mask $M_{s,c}$
\end{algorithmic}
\end{algorithm}

\noindent\textbf{Weight construction}
Given $\sigma_{s,k}^2$, \rl{} constructs a normalized and clipped weight $w_{s,k}$ with a simple normalize–mix–clip rule:
\begin{equation}
\begin{aligned}
w_{s,k}= \mathrm{clip}\!\Big(
(1-\alpha)\,
\frac{\tilde w_{s,k}}{\frac{1}{K}\sum_{j=1}^{K}\tilde w_{s,j}}
+\alpha,\;
w_{\min},\,w_{\max}
\Big), \quad \tilde w_{s,k}=(\sigma_{s,k}^{2}+\varepsilon)^{p},
\end{aligned}
\label{eq:FlowScale-weights}
\end{equation}
where $\varepsilon>0$ avoids numerical issues, $p>0$ controls how strongly weights depend on the noise level, $\alpha\in(0,1)$ mixes the normalized weights with a uniform baseline, and $w_{\min},w_{\max}>0$ bound the effective reweighting.
By construction $\frac{1}{K}\sum_k w_{s,k}=1$ for each $s$, so the average scale of the gradient is preserved.
Because $\sigma_{s,k}$ decreases over time, earlier, noisier steps are relatively upweighted and later, low-noise refinement steps are downweighted, which balances per-step gradient contributions across $k$.
The final $w_{s,k}$ are treated as constants during backpropagation (stop-gradient), so \rl{} only rescales gradients and does not change the optimization target.

In implementation, we broadcast $w_{s,k}$ to shape $[B,S,K,CH,D]$ and multiply it into either the advantages $\hat A_{s,c}$ or the per-step log-probabilities $\log\pi_\theta^{(k)}$ before aggregating over $k$.
These two views are equivalent under the factorization in Eq.~\eqref{eq:factor}.
Alg.~\ref{alg:flowscale} summarizes the computation in PyTorch-style pseudocode.

\subsubsection{Theoretical rationale and derivations}
\label{sec:rationale}
\rl{} is a heuristic reweighting of per-step flow gradients.
Here we sketch a simple rationale based on a Gaussian view of the per-step likelihood and how score norms scale with the noise level~$\sigma_{s,k}$.

\noindent\textbf{Score norm versus noise scale}
Fix an outer step $s$ and flow step $k$, and suppress the indices $(s,c,d)$.
Approximate the per-step likelihood factor $\pi^{(k)}_\theta$ by an isotropic Gaussian in action space with mean $\mu_{s,k}$ and variance $\sigma_{s,k}^2 I$:
\begin{equation}
\log \pi^{(k)}_\theta(a \mid o_s)
= -\,\frac{1}{2\sigma_{s,k}^2}\,\|a-\mu_{s,k}\|^2 + \text{const}.
\label{eq:rationale-gauss}
\end{equation}
We treat $\sigma_{s,k}$ as fixed by the noise schedule (Sec.~\ref{sec:flowscale}) and focus on gradients with respect to the mean $\mu_{s,k}$.
Under this simplification, the score with respect to $\mu_{s,k}$ is:
\begin{equation}
\nabla_{\mu_{s,k}} \log \pi^{(k)}_\theta(a \mid o_s)
= \frac{a-\mu_{s,k}}{\sigma_{s,k}^2}.
\label{eq:rationale-score}
\end{equation}
If we further assume that $a \sim\mathcal{N}(\mu_{s,k},\sigma_{s,k}^2 I)$, then $a-\mu_{s,k}$ has covariance $\sigma_{s,k}^2 I$ and the expected squared norm of the score scales as:
\begin{equation}
\mathbb{E}\big[\|\nabla_{\mu_{s,k}} \log \pi^{(k)}_\theta\|^2\big]
= \mathbb{E}\big[\|a-\mu_{s,k}\|^2\big]\;/\;\sigma_{s,k}^4
\;\propto\; \sigma_{s,k}^{-2}.
\label{eq:rationale-score-norm}
\end{equation}
Thus, in this Gaussian setting, flow steps with smaller noise $\sigma_{s,k}$ tend to produce larger score norms.

\noindent\textbf{Gradient decomposition across flow steps}
Ignoring clipping and KL terms for simplicity, a linearized view of the \fa{} with \rl{} gradient can be written as:
\begin{equation}
\begin{aligned}
\nabla_\theta \mathcal{L}
&\approx -\,\mathbb{E}\!\left[
\sum_{s,c} M_{s,c}\,\hat A_{s,c}
\sum_{k=1}^{K} w_{s,k}\,S^{(k)}_{s,c}
\right], \quad S^{(k)}_{s,c}
\coloneqq \sum_{d=1}^{D}
\nabla_\theta \log \pi^{(k)}_\theta(a_{s,c,d}\mid o_s),
\end{aligned}
\label{eq:rationale-grad}
\end{equation}
where $S^{(k)}_{s,c}$ denotes the contribution of flow step $k$ to the policy gradient at $(s,c)$, and $w_{s,k}$ are the \rl{} weights from Sec.~\ref{sec:flowscale}.
For a fixed $(s,c)$, Eq.~\eqref{eq:rationale-score-norm} suggests that, up to reparameterization:
\begin{equation}
\mathbb{E}\big[\|S^{(k)}_{s,c}\|^2\big] \;\propto\; \sigma_{s,k}^{-2},
\label{eq:rationale-Sk}
\end{equation}
i.e., flow steps with a smaller noise scale $\sigma_{s,k}$ tend to dominate the gradient norm.

\noindent\textbf{A variance-balancing choice of weights}
Motivated by Eq.~\eqref{eq:rationale-Sk}, we consider a variance-balancing criterion: for a fixed $(s,c)$, choose weights $w_{s,k}$ so that the expected contribution of each flow step to the gradient norm is comparable.
Approximating different steps as uncorrelated, we require:
\begin{equation}
\mathbb{E}\big[\|w_{s,k}\,S^{(k)}_{s,c}\|^2\big]
= w_{s,k}^2 \,\mathbb{E}\big[\|S^{(k)}_{s,c}\|^2\big]
\approx \text{constant in }k.
\label{eq:rationale-balance}
\end{equation}
Using Eq.~\eqref{eq:rationale-Sk}, a sufficient choice is:
\begin{equation}
w^\star_{s,k} \;\propto\; \sigma_{s,k},
\label{eq:rationale-wstar}
\end{equation}
so that $w_{s,k}^{\star\,2}\,\sigma_{s,k}^{-2}$ is constant across $k$.
This suggests using weights that grow with the noise scale, downweighting low-noise (high-score) steps and upweighting noisier steps.
In our implementation, we parameterize the weights as
$\tilde w_{s,k} = (\sigma_{s,k}^2+\varepsilon)^{p}$ with $p=0.5$, which gives $\tilde w_{s,k}\propto\sigma_{s,k}$ and is consistent with Eq.~\eqref{eq:rationale-wstar}.

\noindent\textbf{Normalize–mix–clip as a diagonal preconditioner}
The normalize–mix–clip rule in Eq.~\eqref{eq:FlowScale-weights} adds three practical modifications to $w^\star_{s,k}$:
(i) \emph{Normalization} enforces
$\frac{1}{K}\sum_{k} \bar w_{s,k} = 1$ for each $(s,\text{batch})$, so the mean scale of the policy gradient over flow steps is preserved and only relative differences between steps are changed.
(ii) \emph{Uniform mixing} with strength $\alpha$ prevents collapse to a single dominant step by pulling all weights towards one, and
(iii) \emph{Clipping} to $[w_{\min},w_{\max}]$ bounds the effective per-step change in step size.
Together, these operations can be viewed as a simple diagonal preconditioner along the flow-step dimension: \rl{} rescales gradients from different internal steps without changing the overall learning rate or the action-level surrogate in Eq.~\eqref{eq:fa}.
This analysis is only approximate and serves as a motivating heuristic. In practice, the full training dynamics also depend on clipping, KL regularization, and correlations between flow steps.

\subsection{Reward model}
\label{sec:reward}

\noindent\textbf{From RM outputs to advantages}
The reward model (RM) operates at the trajectory level.
Given a rollout $i$ with observations and actions
$\tau^{(i)} = \{o^{(i)}_s, a^{(i)}_{s,c}\}_{s,c}$ and task text $\text{text}_i$, the RM produces a scalar score:
\begin{equation}
R_i = f_{\text{RM}}\big(\tau^{(i)},\,\text{text}_i\big).
\label{eq:rm-score}
\end{equation}
For LIBERO, $R_i\in\{0,1\}$ is a binary success signal; for BRIDGE and real-robot rollouts, $R_i$ comes from a VLM-based classifier.

Following GRPO, we apply group-wise normalization over a batch $\mathcal{G}$ of rollouts:
\begin{equation}
\tilde R_i
= \frac{R_i - \mu_{\mathcal{G}}}{\sigma_{\mathcal{G}} + \varepsilon_R},\quad
\mu_{\mathcal{G}} = \tfrac{1}{|\mathcal{G}|}\sum_{j\in\mathcal{G}} R_j,\quad
\sigma_{\mathcal{G}}^2 = \tfrac{1}{|\mathcal{G}|}\sum_{j\in\mathcal{G}} (R_j-\mu_{\mathcal{G}})^2,
\label{eq:rm-normalized}
\end{equation}
with a small constant $\varepsilon_R>0$ for numerical stability.
The normalized score $\tilde R_i$ is broadcast to all chunks $(s,c)$ within trajectory $i$ and used as the advantage:
\begin{equation}
\hat A^{(i)}_{s,c} = \tilde R_i\,M^{(i)}_{s,c},
\label{eq:rm-adv}
\end{equation}
where $M^{(i)}_{s,c}$ is the padding/termination mask from Sec.~\ref{sec:rl_setup}.
These advantages are plugged into the \fa{} and \rl{} objectives
(Eqs.~\eqref{eq:fa} and \eqref{eq:flowscale}) in place of the scalar $\hat A_t$ in GRPO.

\subsubsection{RMs for LIBERO}
For LIBERO, we can roll out policies in the physics simulator and obtain ground truth success labels and episode lengths.
We collect rollouts with a horizon of 500 frames per episode, render each rollout as a video, and record a binary success label and the number of environment steps until success.
Using these labeled videos, we fine-tune a Qwen2.5-VL-7B~\cite{bai2025qwen2} as a binary RM: given a task description and frames resized to $224\times224$, the model predicts whether the task is completed (and an estimated completion step for temporal masking).
At inference, we uniformly subsample 50 frames per trajectory, evaluate the RM five times with stochastic decoding via vLLM~\cite{kwon2023efficient}, and take the majority vote as the final label.
This binary outcome is mapped to $R_i$ in Eq.~\eqref{eq:rm-score} for LIBERO policies trained both in the simulator and in the world model.

Directly applying a simulator-trained RM to world-model rollouts is suboptimal, since world-model videos differ in appearance and long-horizon dynamics.
We therefore adopt a simple domain-bridging strategy: during data collection, we run the same policy in both the simulator and the world model in parallel with identical action sequences.
The simulator provides ground truth success labels, while the world model provides the RGB observations that the RM will see during training and RL.
We train a second RM to predict simulator-derived labels from world-model videos, yielding an RM that is visually adapted to the world model but supervised by the reliable simulator signal.
In practice, this semi-synthetic supervision is a pragmatic workaround that we found adequate to run RL in the world model, although the induced rewards remain noisy and the stability of the resulting training signal varies across tasks.

\subsubsection{RMs for BRIDGE and real-world data}
\begin{figure}[t]
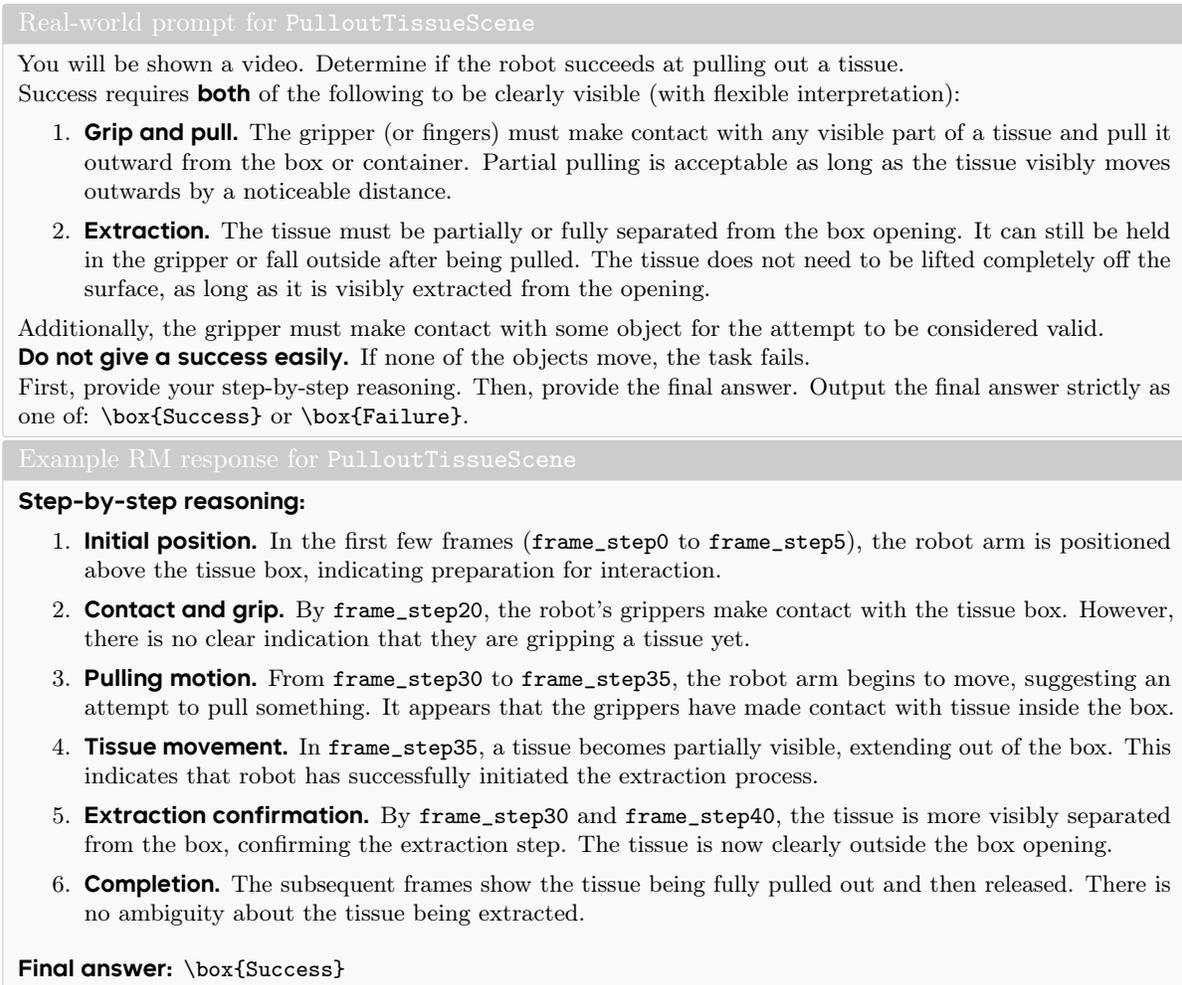

  \centering
  \begin{subfigure}[t]{0.95\linewidth}
    \begin{promptbox}{Real-world prompt for \texttt{PulloutTissueScene}}
    \small

    You will be shown a video. Determine if the robot succeeds at pulling out a tissue.

    Success requires \textbf{both} of the following to be clearly visible (with flexible interpretation):

    \begin{enumerate}
      \item \textbf{Grip and pull.}
      The gripper (or fingers) must make contact with any visible part of a tissue and pull it outward from the box or container. Partial pulling is acceptable as long as the tissue visibly moves outwards by a noticeable distance.

      \item \textbf{Extraction.}
      The tissue must be partially or fully separated from the box opening. It can still be held in the gripper or fall outside after being
      pulled. The tissue does not need to be lifted completely off the surface, as long as it is visibly extracted from the opening.
    \end{enumerate}

    Additionally, the gripper must make contact with some object for the attempt to be considered valid.

    \textbf{Do not give a success easily.} If none of the objects move, the task
    fails.

    First, provide your step-by-step reasoning. Then, provide the final answer. Output the final answer strictly as one of:
    \verb|\box{Success}| or \verb|\box{Failure}|.
    \end{promptbox}
  \end{subfigure}

  \begin{subfigure}[t]{0.95\linewidth}
    \begin{promptbox}{Example RM response for \texttt{PulloutTissueScene}}
    \small

    \textbf{Step-by-step reasoning:}

    \begin{enumerate}
      \item \textbf{Initial position.}
      In the first few frames (\verb|frame_step0| to \verb|frame_step5|), the robot arm is positioned above the tissue box, indicating preparation for interaction.

      \item \textbf{Contact and grip.}
      By \verb|frame_step20|, the robot's grippers make contact with the tissue box. However, there is no clear indication that they are gripping a tissue yet.

      \item \textbf{Pulling motion.}
      From \verb|frame_step30| to \verb|frame_step35|, the robot arm begins to move, suggesting an attempt to pull something. It appears that the grippers have made contact with the tissue inside the box.

      \item \textbf{Tissue movement.}
      In \verb|frame_step35|, a tissue becomes partially visible, extending out of the box. This indicates that the robot has successfully initiated the extraction process.

      \item \textbf{Extraction confirmation.}
      By \verb|frame_step30| and \verb|frame_step40|, the tissue is more visibly separated from the box, confirming the extraction step. The tissue is now clearly outside the box opening.

      \item \textbf{Completion.}
      The subsequent frames show the tissue being fully pulled out and then released. There is no ambiguity about the tissue being extracted.
    \end{enumerate}

    \medskip
    \textbf{Final answer:} \verb|\box{Success}|

    \end{promptbox}
  \end{subfigure}
  \caption{
    Real-world reward model prompt and example response for the task \texttt{PulloutTissueScene} on BRIDGE.
    The prompt follows the principles in Sec.~\ref{sec:rm_discussion} (high recall with reasonably clean positive labels). Although it is not optimal, it provides sufficiently informative supervision for our current real-world experiments.
}
  \label{fig:rm_prompt_response_tissue}
\end{figure}

For BRIDGE tasks evaluated on SimplerEnv-WidowX~\cite{li2025evaluating} and for our UR30e real-robot experiments, collecting diverse labeled failures is expensive and off-policy logs rarely contain clean success/failure annotations.
In these settings we use a larger off-the-shelf VLM as a zero-shot RM.
Concretely, we deploy Qwen2.5-VL-72B as a real-world RM: for each trajectory, we sample 20 frames, resize them to $224\times224$, and feed them together with the task description and a task-specific prompt into the model via vLLM.
An example prompt and corresponding RM response for \texttt{PulloutTissueScene} are shown in Fig.~\ref{fig:rm_prompt_response_tissue}.
Trajectories are short (30 steps on BRIDGE and 100 steps on UR30e), so 20 frames suffice to cover the motion.
We again use a voting scheme over five evaluations and take the majority decision as the final label $R_i$.

Unlike the LIBERO RM, which also predicts an estimated finish step used for temporal masking, the real-world RM only outputs a binary success or failure label.
Because rollouts are short, we use an all-ones temporal mask in Eq.~\eqref{eq:rm-adv}, which empirically suffices for effective improvement.

%% file: sections/4-experiments.tex
\section{Experiments}

\subsection{Experimental setups}

\subsubsection{\wm{} training setups} 
Our \wm{} is initialized from Cosmos-Predict2-2B-Video2World~\cite{agarwal2025cosmos} and augmented with history-aware mechanism, and dual action conditioning. 
\wm{} conditions on the first observed frame and, given a 20-step action chunk, generates the next 20 frames.
For action frame construction, we set $l=0.15$, $\mathbf{r}_{\text{ref}}=40$, $z_{\text{ref}}=1.0$, $\mathbf{r}_\text{min}=8$, and $\mathbf{r}_{\text{max}}=140$.
Across all datasets, we normalize the gripper signal to $[g_{\min}, g_{\max}] = [0, 1]$, where $0$ denotes fully closed and $1$ denotes fully open.
For the video autoencoder $\mathrm{E}$, we adopt the Wan2.1 video autoencoder~\cite{wan2025} (also used in Cosmos-Predict2~\cite{agarwal2025cosmos}), which compresses the spatio-temporal dimensions of a video by $4 \times 8 \times 8$, yielding latents of size $H_l = H / 8$, $W_l = W / 8$, $T_l = 1 + T / 4$, and $C_l = 16$.
For our \wm{}, the total number of parameters is 2.058B, the DiT channel $D_{m} = 2048$.
For the history-aware mechanism, we set $T_h = 60$, i.e., we maintain a fixed-length buffer of the most recent 60 frames as historical latent input.

We pretrain the \wm{} on a mixture of robot manipulation datasets, including AgiBot~\cite{bu2025AgiBot}, DROID~\cite{khazatsky2024droid}, LIBERO~\cite{liu2023libero}, and high-quality subsets filtered from Open-X~\cite{o2024open}, with a total of over 31M sampled trajectories. 
Since not all Open-X sub-datasets are suitable for our setting (some videos have extremely low resolution, some robots have poor end-effectors, and some subsets do not provide reliable end-effector poses or gripper states), we only use a curated subset for pretraining. 
Concretely, we select Austin Sailor~\cite{nasiriany2022sailor}, DLR Wheelchair Shared Control~\cite{vogel_edan_2020, quere_shared_2020}, BC-Z~\cite{jang2022bc}, CMU Stretch~\cite{bahl2023affordances, mendonca2023structured}, Stanford HYDRA~\cite{belkhale2023hydra}, USC Jaco Play~\cite{dass2023jacoplay}, Furniture Bench~\cite{heo2023furniturebench}, NYU Franka Play~\cite{cui2022play}, and RT-1-style data~\cite{brohan2022rt}. 
For downstream RL, we fine-tune \wm{} on BRIDGE~\cite{walke2023bridgedata}, LIBERO~\cite{liu2023libero}, and our self-collected real-robot dataset, all of which provide simulation or real-robot environments for evaluation.

Pretraining and fine-tuning data are strictly separated. 
For pretraining, we train all model parameters on 64 H200 GPUs for 2 epochs, with a batch size of 16 per GPU and gradient accumulation of 4. For fine-tuning, we directly optimize all model parameters on 8 H200 GPUs with a batch size of 24, running for a task-dependent number of steps.
Both stages use the fused Adam optimizer with a learning rate of $1\times10^{-4}$ and weight decay of $0.1$.

During both \wm{} pretraining and fine-tuning, we keep dataset-specific input resolutions. 
On AgiBot~\cite{bu2025AgiBot} and our custom real-robot data we use $240\times320$, on DROID~\cite{khazatsky2024droid} we use $240\times416$, and on LIBERO~\cite{liu2023libero} and BRIDGE~\cite{walke2023bridgedata} we use $256\times256$. 
For Open-X, we also standardize to $240\times320$: we first resize the height to 240 pixels, then either pad black borders on the left and right if the width is smaller than 320, or center-crop to 320 if the width is larger. 
Since pretraining spans multiple sources and only some datasets provide the camera parameters required to construct action frames, our batch sampler always draws each mini-batch from a single dataset. 
This avoids conflicts in resolution and conditioning signals and stabilizes multi-dataset pretraining.

\subsubsection{Real-world experiment setting}
\label{sec:real_world_setting}
We collected our custom manipulation dataset using a UR30e robot arm, providing a physical evaluation environment to assess real-world adaptation of \wm{} and the universality of our RL algorithm.
We collected 800 trajectories from 4 tasks.
For all tasks, we fix the camera, low-level controller, and policy interfaces, and vary only the initial object configurations according to a predefined grid layout on the table.
For each policy–task pair, we conduct three evaluation runs and report the mean and standard deviation of the success rate across all trials.

To stabilize RL on the real-robot, we augment the visual input with a low-dimensional state vector.
For each rollout, the policy receives the first-frame RGB image together with the initial robot state (end-effector pose and gripper status).
The action head predicts delta actions, and we update the state for subsequent action inputs by integrating these deltas over time rather than querying the robot at every step.
Thus, in all real-world RL experiments, the training data consists of a single initial image and the state trajectory induced by the predicted delta actions.

\begin{figure*}
    \includegraphics[width=1.0\linewidth]{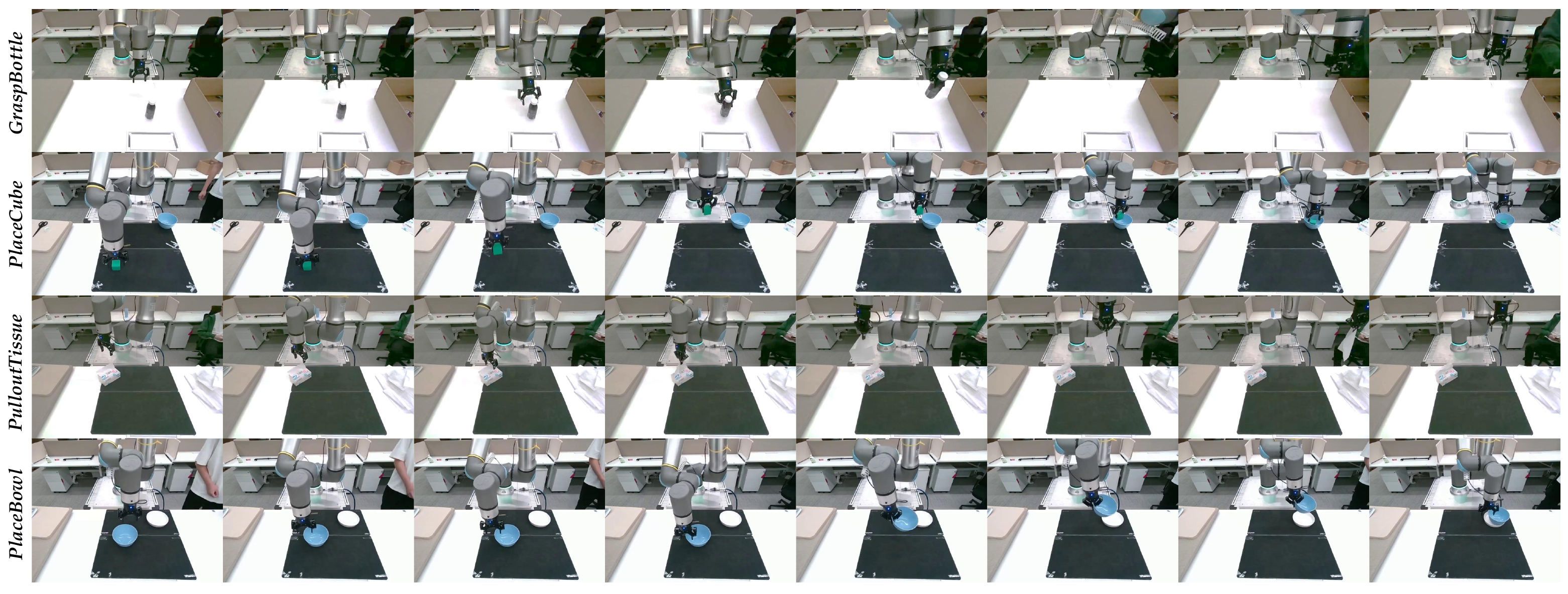}
    \caption{\textbf{Presentation of custom data collected using a UR30e robot arm.} We collect data for four tabletop manipulation tasks, including challenging cases such as pulling tissues from a box, which are impossible to simulate accurately in standard physics simulators.}
    \label{fig:real-robot}
\end{figure*}

\noindent\textbf{Data collection}
Real-world data for these tasks are collected using a UR30e robot arm teleoperated through the GELLO~\cite{wu2024gello} interface.
A fixed third-person Intel RealSense D455 RGB-D camera provides visual observations, and its extrinsics are calibrated to the robot base using the EasyHandEye toolkit.
During data collection, objects are uniformly placed across the workspace to capture diverse initial configurations, and each demonstration records a complete manipulation attempt from the first frame to the end of the episode.
Data from four tasks are collected, i.e., \emph{GraspBottle}, \emph{PlaceCube}, \emph{PulloutTissue}, and \emph{PlaceBowl}.

\noindent\textbf{Details of each task}
Fig.~\ref{fig:real-robot} shows a random subset of trajectories collected on each task:

\noindent\textbf{(i)} \emph{GraspBottle}
A plastic bottle is placed on the table and the goal is to pick it up and place it into a box.
During data collection, the bottle is uniformly placed at different locations on the tabletop.
The bottle is smooth and slightly elastic, so even small grasp tilts can cause it to slip or pop out of the gripper, making the task very sensitive to grasp pose and stability.
For evaluation, we draw a regular \(4\times 5\) grid on the table and place the bottle once at each grid cell, yielding 20 distinct start positions per run.

\noindent\textbf{(ii)} \emph{PlaceCube}
The policy must place a green cubic block into a bowl. 
The cube starts from a fixed location on the table, while the bowl position is varied.
At evaluation time, we use the same \(4\times 5\) grid layout: the cube remains at its fixed start pose, and the bowl is placed at each grid cell in turn, resulting in 20 trials per run that cover the full grid.

\noindent\textbf{(iii)} \emph{PulloutTissue}
The policy must grasp a tissue from a box, pull it out, and place it next to the box.
Because the tissue is soft and deformable, the gripper must align accurately with the exposed edge and pull smoothly without tearing or dropping it.
For evaluation, we place the box at each cell of a \(4\times 5\) tabletop grid, yielding 20 trials per run.

\noindent\textbf{(iv)} \emph{PlaceBowl}
The policy must pick up a bowl and place it onto a plate.
The plate remains fixed while the bowl position is varied.
We define a \(2\times 5\) grid of bowl start positions and evaluate two full passes over this grid per run, resulting in 20 trials per run.
Compared with \emph{PlaceCube}, this task emphasizes stable bowl grasps and precise placement on a relatively small plate support area.

Across all four tasks, each method is evaluated over three runs, so reported numbers are the mean and standard deviation of the success rate over \(3\times 20\) trials per task.

\subsubsection{Supervised fine-tuning (SFT) policies and RL setups}
We evaluate three policies at different scales, i.e., VLA-Adapter-0.5B~\cite{wang2025vla}, Pi0.5-3B~\cite{intelligence2504pi0}, and OpenVLA-OFT-7B~\cite{kim2025openvla}.
All policies take a single-image per step and output a 7D delta action via a lightweight flow action head.
Before RL, we conduct fine-tuning for the policies, with batch size 64, a learning rate of $2.5e^{-5}$ with AdamW optimizer and weight decay 0.1 for 200k steps.
For real-robot experiments, we independently fine-tune policies on each of the four tasks (200 trajectories per task) with batch size 16 for 50k steps per task using the same optimizer.
For RL, we set $\text{ngroup} = 8$, a total batch size of 256, and a mini-batch size of 128, with all policies trained for 100 steps.
All RL-related experiments are conducted on 8 H200 GPUs.

\FloatBarrier
\subsection{World model evaluation}

\begin{figure*}
    \centering
    \includegraphics[width=1.0\linewidth]{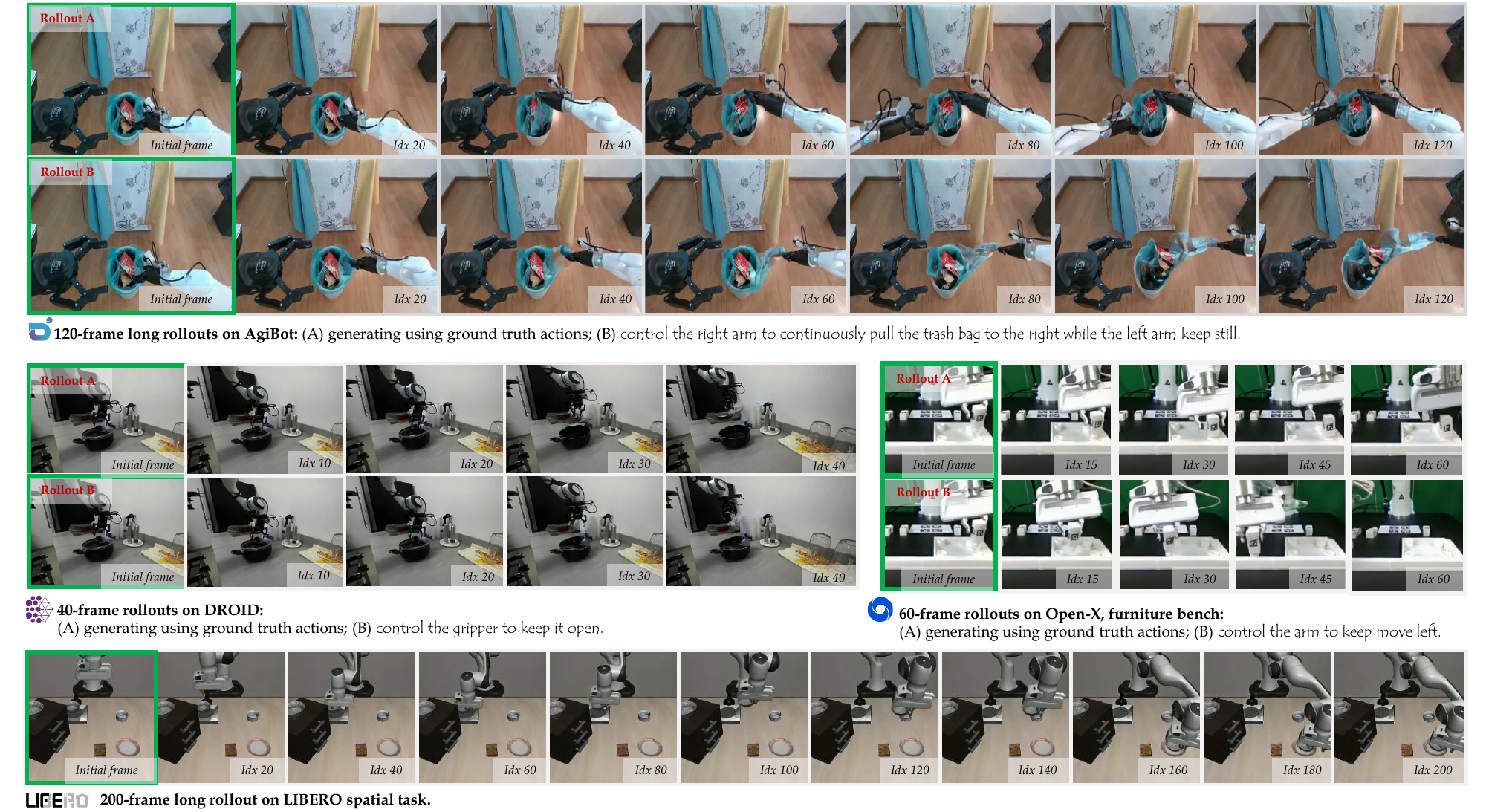}
    \caption{\textbf{Qualitative results of the pretrained \wm{}.} Our \wm{} accurately maps actions to robot manipulation, supports hundred-level-frame rollouts, and enables flexible control of both arm motions and end-effector states. Even for unseen action controls at training time, generations remain accurate and physically plausible.}
    \label{fig:pretrain_viz}
\end{figure*}

\subsubsection{Evaluation of the pretrained \wm{}}
Before pretraining, we unify data conventions across corpora, e.g., standardizing gripper state semantics, coordinate frames (camera, robot, and world), and action parameterizations, to eliminate cross-dataset ambiguities and enable efficient large-scale joint training. 
In particular, since data in Open-X lacks reliable camera parameters, it is conditioned only by the scalar action stream, whereas all other datasets additionally employ action frame conditioning.

\input{tables/pretraining_results}

Tab.~\ref{tab:pretrain} presents detailed validation results of the pretrained \wm{} on held-out trajectories from each dataset. 
Across multiple datasets, the model attains consistently high visual fidelity and strong action consistency.
Note that these numbers are obtained from a single pretrained \wm{}.
With dataset-specific fine-tuning the metrics improve substantially, as evidenced by the large gains on LIBERO reported in Tab.~\ref{tab:libero_ablation}.

\begin{figure*}
    \centering
    \includegraphics[width=1.0\linewidth]{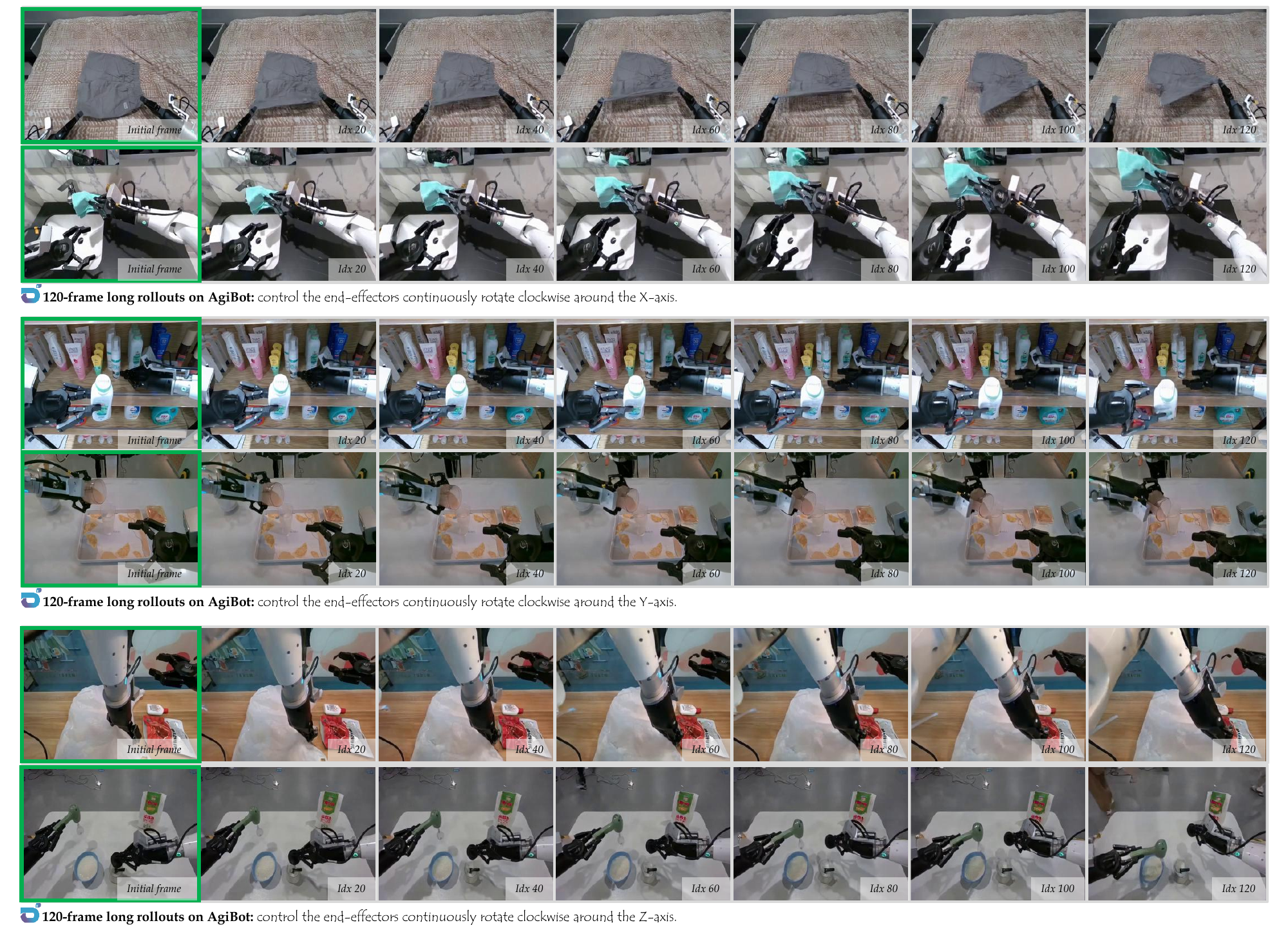}
    \caption{\textbf{Qualitative results of the pretrained \wm{} on AgiBot.} We additionally visualize rollouts where the end-effector is commanded to perform mechanical rotations and physically interact with the real-world. Our \wm{} produces highly realistic and physically consistent results, including accurate shadows, specular reflections, liquid pouring, and interactions with complex materials.}
    \label{fig:pretrain_viz_supp_agibot}
\end{figure*}

Representative rollouts in Fig.~\ref{fig:pretrain_viz} illustrate long-horizon end-effector control, contact formation, and object state changes under varied viewpoints.
Fig.~\ref{fig:pretrain_viz_supp_agibot} shows more qualitative results of the pretrained \wm{} on AgiBot.
These examples are drawn from held-out validation trajectories, and the specific man-made motion patterns shown here never appear in the pretraining data.
Nevertheless, the pretrained \wm{} already captures rich, physically plausible interactions with the environment across diverse scenes and object configurations.
It not only preserves fine-grained appearance details, but also reproduces realistic contact dynamics and secondary effects (e.g., shadows, reflections, and deformations of soft or thin objects), closely matching real-world physics even under novel combinations of actions and objects.
Such faithful physical behavior on unseen sequences is crucial for providing reliable rollouts when optimizing VLA policies via RL in the real world, since the policy is trained almost entirely on model-generated experience.

\subsubsection{Fine-tuning on BRIDGE}
To further assess generalization of \wm{} under limited or rich real-robot data, we fine-tune on BRIDGE~\cite{walke2023bridgedata} in three settings summarized in Tab.~\ref{tab:bridge_finetune}. 
In experiment 1, we select tasks such as `fold the cloth', fine-tune using only a small snippet of the available trajectories per task, and evaluate generalization on the remaining held-out trajectories and scene instances.
In experiment 2, we increase difficulty by testing cross-object transfer under few-shot supervision. 
For example, we fine-tune on 150 demonstrations of a `pick up sth.' (exclude carrot) task and evaluate zero-shot on the complementary `pick up carrot' task.
In experiment 3, we examine the upper bound with data-rich fine-tuning.
We allocate $15/16$ of the trajectories for training and the remaining $1/16$ for validation, preserving scene disjointness across splits.
We compare our \wm{} with state-of-the-art baselines~\cite{agarwal2025cosmos, HaCohen2024LTXVideo, liao2025genie, blattmann2023stable, zhu2024irasim, guo2025ctrl}, and all baselines are well fine-tuned using the same action-conditioning strategy with \wm{}.
In experiments 1 and 2, each model is fine-tuned for 2k steps, while in experiment 3 we extend fine-tuning to 30k steps. 
For all these experiments, action frame conditioning is disabled for our \wm{}.

\begin{figure*}
    \centering
    \includegraphics[width=1.0\linewidth]{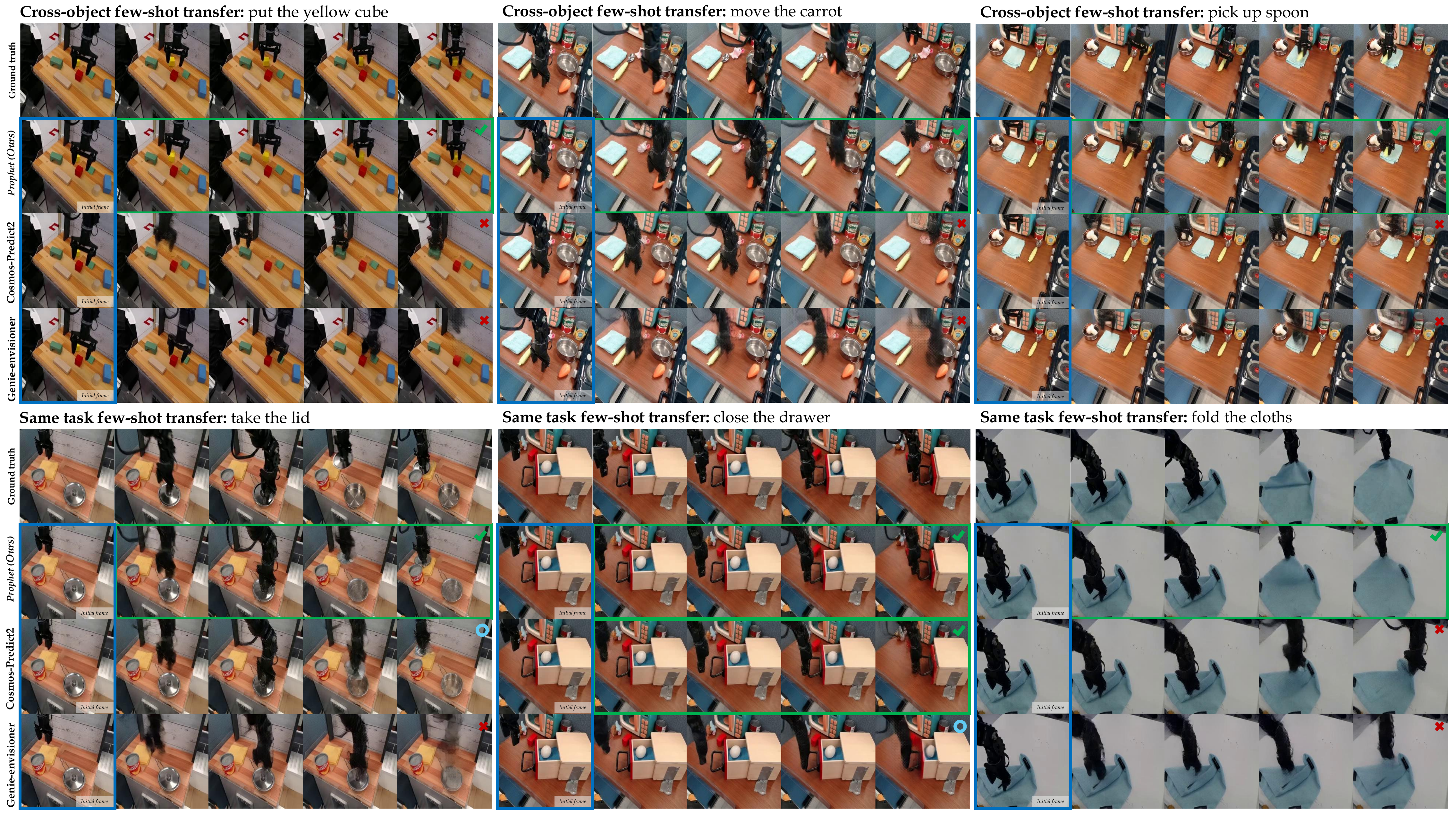}
    \caption{\textbf{Qualitative few-shot transfer results on BRIDGE.} We use a green \checkmark~to denote trajectories that correctly complete the task, a blue \(\bigcirc\) for cases that largely succeed but exhibit noticeable execution deviations, and a red \(\times\) for rollouts that fail to perform the intended task or do not produce a recognizable outcome. Our \wm{} achieves perfect success on all six challenging few-shot transfer tasks, with action execution and physical interactions that are nearly indistinguishable from the real videos. In contrast, the baselines perform poorly under this setting, often grasping the wrong object, following incorrect trajectories, or failing to generate meaningful rollouts at all.}
    \label{fig:finetune_viz_supp_bridge}
\end{figure*}

\input{tables/bridge_finetuning}

Quantitative results are shown in Tab.~\ref{tab:bridge_finetune}.
Our \wm{} consistently matches or exceeds baselines on visual fidelity while yielding markedly better action consistency.
The gains are most pronounced in the few-shot regimes of experiments 1 and 2, indicating that \wm{} has already learned an action-to-manipulation paradigm during pretraining, enabling rapid generalization from only small amounts of new data.
In experiment 3, even with abundant data, \wm{} exhibits across-the-board advantages, highlighting its strong generality and faithful modeling of end-effector trajectories and contacts.

In Fig.~\ref{fig:finetune_viz_supp_bridge}, we show qualitative comparison samples from the few-shot transfer experiments on BRIDGE, encompassing all six few-shot settings.
In these experiments, all world models are given the same ground truth action sequences as conditioning, the only difference is how faithfully they realize these actions in the generated videos. 
Our \wm{} generalizes well from the few-shot fine-tuning data, it consistently follows the conditioned actions, closes the correct drawer, grasps the intended object, and produces physically plausible interactions on unseen scenes. 
By contrast, the baselines often ignore the action condition, producing rollouts that grasp the wrong object, follow an incorrect trajectory, or fail to produce a recognizable outcome.
These failures indicate weak alignment between their latent dynamics and the action channel.
This strong action-faithful generalization from few-shot world model fine-tuning is precisely what enables the subsequent RL stage on BRIDGE, where both full-data and few-shot policy optimization benefit significantly from rollouts that accurately reflect how different action sequences affect the physical scene.

\begin{figure*}
    \includegraphics[width=1.0\linewidth]{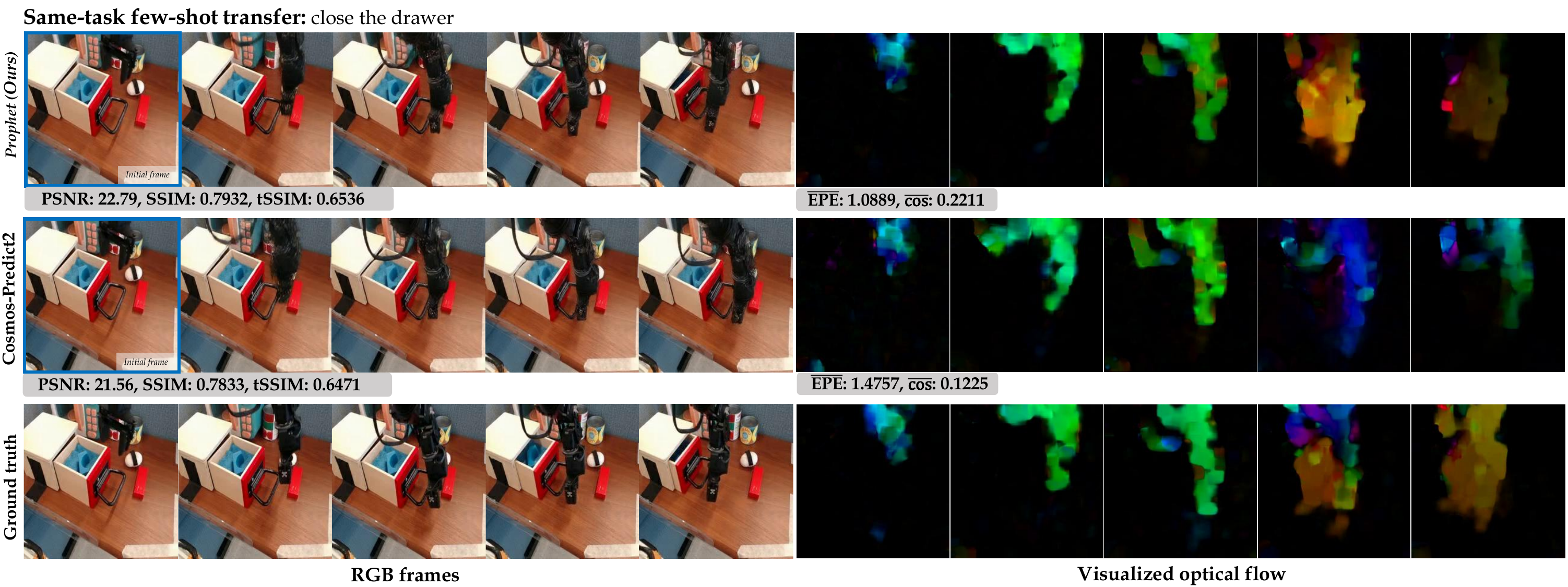}
    \caption{\textbf{Optical flow evaluation reveals action errors beyond visual metrics.} Although the baseline clearly fails to fully execute the closing motion, standard visual fidelity metrics differ only slightly between the two models. On the right, the visualized optical flow focuses on the moving arm and drawer. The flow of our \wm{} closely matches the ground truth, whereas the baseline exhibits obvious mismatches in motion and contact, leading to a much larger gap in flow-based metrics that cleanly separates good from poor rollouts.}
    \label{fig:optical_flow}
\end{figure*}

Notably, several baselines report relatively high PSNR and SSIM yet fail to execute the intended action.
We use a combination of qualitative and quantitative analysis to assess why optical flow-guided evaluation provides more effective metrics for action-conditioned world models, shown in Fig.~\ref{fig:optical_flow}.

Conventional frame-level metrics such as PSNR and SSIM primarily measure global visual fidelity; they are sensitive to blur, noise, and global color or texture shifts, but largely agnostic to whether the `motion' in the generated video is behaviorally correct.
In manipulation settings, however, two rollouts may look equally sharp and photorealistic while corresponding to very different actions (e.g., the drawer only half closed or the gripper stopping short of contact). 
As illustrated on the left of Fig.~\ref{fig:optical_flow}, both our \wm{} and the baseline produce high visual fidelity, leading to only marginal differences in visual fidelity despite the baseline clearly failing to complete the `close the drawer'.

To make the evaluation more sensitive to action correctness, we compute dense optical flow between consecutive frames for both the generated rollout and the ground truth, and compare them only in regions with significant motion. 
Our flow-based metrics are largely invariant to static background appearance and lighting, but are highly sensitive to whether the arm moves along the correct trajectory and establishes contact at the right time and location.
On the right of Fig.~\ref{fig:optical_flow}, the visualized flow fields highlight this effect: the flow of our \wm{} closely tracks the ground truth motion of both the arm and the drawer, whereas the baseline exhibits fragmented or misplaced motion, especially near the handle and along the closing direction. 
Correspondingly, the gap in EPE and flow-direction cosine between the two models is much larger than the gap in PSNR and SSIM. 
Across tasks, we observe that these optical flow metrics correlate much more strongly with task success rates of downstream VLA policies, indicating that flow-guided evaluation provides a more faithful measure of action-conditioned world model quality than conventional metrics.

\subsubsection{Fine-tuning on custom data}
We fine-tuned \wm{} for 20k steps on custom data.
Tab.~\ref{tab:sii_finetune} shows the results on validation samples.
Our \wm{} achieves strong generation quality, with each introduced component yielding considerable gains.

\input{tables/sii_finetuning}

\begin{figure*}
    \centering
    \includegraphics[width=1.0\linewidth]{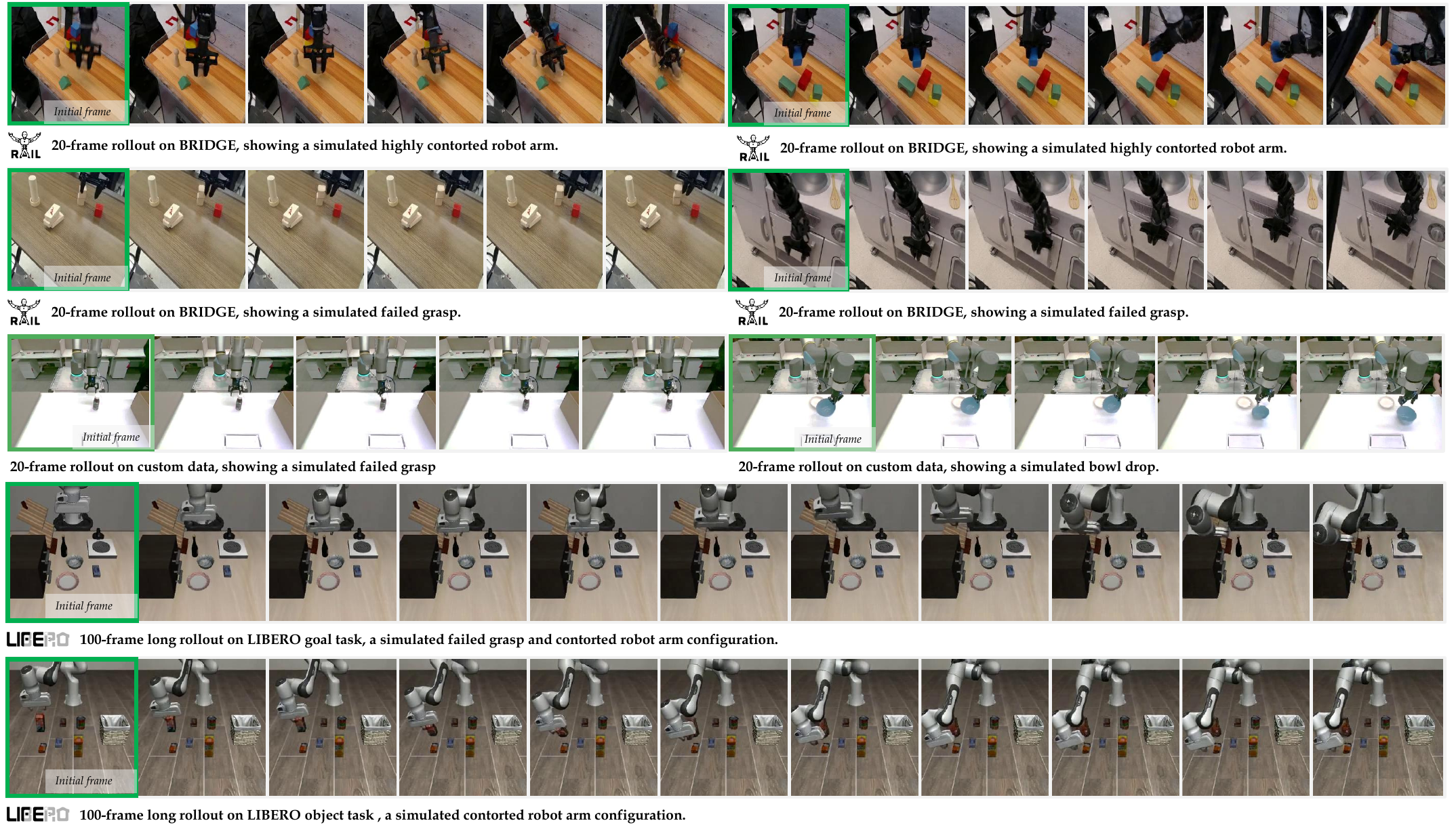}
    \caption{\textbf{Simulated failure rollouts by the fine-tuned \wm{} on each dataset.} We visualize diverse simulated failure rollouts, including failed grasps, highly contorted arm poses, and interactions with task-irrelevant objects. Crucially, the \wm{} not only generates successful trajectories but also realistic, precise failures, allowing downstream RL to better optimize policies by learning from both success and failure.}
    \label{fig:simulated_failures}
\end{figure*}

\input{tables/libero_finetuning}

\subsubsection{Ablation studies on \wm{} components}
We conduct additional ablation studies of key \wm{} components on LIBERO.
We fine-tune \wm{} for 30k steps using the original training rollouts augmented with trajectories generated by SFT VLA-Adapter-0.5B, and evaluate on unseen trajectories. 
As shown in Tab.~\ref{tab:libero_ablation}, using pretraining and the history-aware mechanism yields consistent, across-the-board gains.
In the LIBERO scenario, introducing the action frame further boosts action consistency, but incurs a small drop in PSNR.
Given our downstream goal, i.e., faithful action execution, this trade-off is desirable.

\subsubsection{Simulating failure rollouts}

A key advantage of using a world model for RL is that it can expose the policy to both successes and failures without repeatedly failing on real hardware. Rather than only replaying curated demonstrations, we would like \wm{} to also hallucinate plausible but undesired outcomes, so that the policy can practice avoiding unsafe or unproductive behaviors entirely inside the model.

Fig.~\ref{fig:simulated_failures} shows simulated failure rollouts generated by the fine-tuned \wm{} on three domains. 
On BRIDGE data, even though \wm{} is fine-tuned only on successful demonstrations using full data, it produces realistic failures such as stopping short of the target, or drifting away after contact, effectively enriching the data distribution seen by downstream RL. 
On our custom real-robot data, even a small data snippet is sufficient for adapting \wm{} to produce plausible rollouts under challenging action sequences, suggesting that it can serve as a useful learned simulator under very limited supervision.
On LIBERO data, the fine-tuned \wm{} captures how small perturbations in the action sequence lead to missed grasps, contorted arm poses, or spurious interactions with distractor objects, rather than collapsing to only ideal outcomes.

Across these settings, the ability to generate calibrated failure rollouts is crucial for RL, since it provides informative negative examples, teaches the policy which behaviors to avoid, and prevents overfitting to overly optimistic dynamics that would rarely occur in the real environment.

\subsection{RL with world models across simulators and real-robot}
\label{sec:wm_rl_eval}

\input{tables/simplerenv}
\input{tables/rl_baseline_comparison}
\input{tables/simplerenv_windowx_joint_ablation}
\input{tables/simplerenv_wm_ablation}
\input{tables/simplerenv_windowx_num_demo_ablation}
\stepcounter{table}
\FloatBarrier

\subsubsection{Evaluation on BRIDGE and SimplerEnv}
\noindent\textbf{Single-task RL with \wm{}}
We use \wm{} fine-tuned on BRIDGE full data for 30k steps.
Policies are first SFT for 200k steps on the same BRIDGE data, and then post-trained in \wm{} with \fa{} and \rl{}.
We consider four WidowX tasks: \emph{PutCarrot}, \emph{PutSpoon}, \emph{StackCube}, and \emph{PutEggplant}.
For each task, we run RL separately, initializing the policy and world model with only the language instruction and 100 single-image snapshots, and then training the policy with our paradigm (Fig.~\ref{fig:pipe}).
Performance is reported on the corresponding SimplerEnv benchmark.

Tab.~\ref{tab:simplerenv_widowx} reports grasp and success rates.
Across all three backbone sizes with the same flow head, both metrics improve consistently after RL.
\wm{} is fine-tuned on real BRIDGE data, whereas SimplerEnv, although sharing a similar tabletop setup, differs in visuals and object instances.
The gains therefore reflect transfer from real-world data to a distinct simulator.
Using 100 single-image snapshots per task for post-training yields improvements, indicating that \wm{} can provide accurate rollouts for policy optimization in this setting and that our RL procedure improves VLA policies from modest real-world supervision.
For additional RL baselines, Tab.~\ref{tab:rl_baseline_comparison} compares success gains when post-training the Pi0.5 SFT policy from Tab.~\ref{tab:simplerenv_widowx}.
We use official VLA-RFT/WMPO numbers when available; for RIPT-VLA, which assumes environment rollouts and binary rewards, we use its RLOO/PPO objective on the same \wm{}/RM backend.
This RIPT-VLA-style variant tests the objective under a matched learned-rollout interface, and its 12.4 gain is close to but below our 13.8 on \textsc{PutCarrot}; on every task and in the overall average, all baselines remain below our method, suggesting that \fa{} and \rl{} add optimizer-side gains beyond the shared backend.

\noindent\textbf{Iterative RL with low-reward scene initialization.}
We evaluate iterative RL on the PutCarrot task using Pi0.5.
Each round continues from the previous policy checkpoint, while low-reward scene configurations are used to initialize \wm{} rollouts for the next round.
As shown in Tab.~\ref{tab:iterative_rl_carrot}, success increases from 29.2 after SFT to 43.0, 45.8, and 54.2 over three RL rounds.
This yields a total gain of 25.0 points, showing that focusing later rounds on the remaining failure cases enables continued policy improvement.

\noindent\textbf{Multi-task RL with \wm{}}
To test whether our training paradigm also works in a multi-task setting, we run an additional experiment on the four SimplerEnv-WidowX tasks and jointly fine-tune each VLA variant on the union of these tasks in \wm{}.
We initialize from the same 200k-step SFT checkpoints as in Tab.~\ref{tab:simplerenv_widowx}, then perform 250 RL updates with \fa{} and \rl{}, sampling the four tasks uniformly when generating world model rollouts; evaluation remains per-task in the simulator.
Tab.~\ref{tab:simplerenv_windowx_joint_ablation} shows that multi-task RL with \wm{} yields consistent gains across all three variants, improving both grasp metrics and full-task success for each individual task.
This suggests that \fa{} with \rl{} can effectively exploit a shared world model when multiple tasks are optimized jointly, without requiring task-specific RL runs.

\noindent\textbf{World model choice and fine-tuning data}
Tab.~\ref{tab:simplerenv_windowx_wm_ablation} studies how the choice of world model and the amount of fine-tuning data influence RL performance.
All experiments use single-image VLAs, with \fa{} and \rl{} for policy optimization.
We fine-tune Cosmos-Predict2 and \wm{} on BRIDGE full data for 30k steps and use each as the rollout generator for the same policies under identical RL settings.
Replacing Cosmos-Predict2 with \wm{} increases success rates, suggesting that a more accurate action-conditioned world model yields stronger downstream VLA performance.
In a few-shot setting, we fine-tune both models on only 400 BRIDGE samples for 500 steps.
Here the Cosmos-Predict2 setup degrades more, whereas \wm{} largely preserves its gains, indicating better sample efficiency and few-shot adaptability as an RL backend.

\begin{figure}[!htbp]
    \centering
    \includegraphics[width=0.66\textwidth]{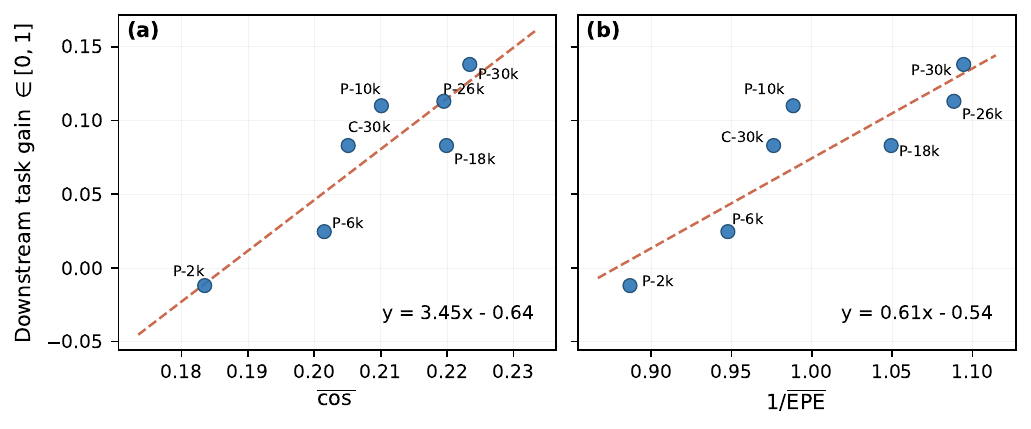}
    \caption{\textbf{Correlation between world model metrics and RL task gains.} \textbf{P/C-$x$k} denotes \wm{} and Cosmos-Predict2 fine-tuned at $x$k steps on BRIDGE.}
    \label{fig:corr}
\end{figure}

\noindent\textbf{World model metrics and RL gains}
We test whether action-faithful prediction quality explains downstream policy improvement.
For \textit{PutCarrot} in SimplerEnv, we evaluate world model checkpoints with a fixed Pi0.5 policy and relate the resulting task gain to optical-flow metrics.
As shown in Fig.~\ref{fig:corr}, both $\overline{\cos}$ and $1/\overline{\mathrm{EPE}}$ correlate strongly with gains across checkpoints: Spearman $\rho=0.8469$ ($p=0.0162$) for $\overline{\cos}$ and $\rho=0.9370$ ($p=0.0019$) for $1/\overline{\mathrm{EPE}}$ (Kendall $\tau=0.7807/0.8783$).
This indicates that more action-faithful rollouts tend to yield larger downstream gains, making flow-based metrics a useful proxy for checkpoint selection.

\noindent\textbf{Few-shot RL with \wm{}}
We study how sensitive our training paradigm is to the amount of data to bootstrap RL with \wm{}.
We use VLA-Adapter-0.5B~\cite{wang2025vla} on four SimplerEnv-WidowX tasks.
Starting from the 200k-step SFT checkpoint in Tab.~\ref{tab:simplerenv_widowx}, we subsample each task to either 100 or 10 images, which serve as the observations for initializing RL with \wm{}.
We post-train with \fa{} and \rl{} for 100 updates.

Tab.~\ref{tab:simplerenv_windowx_num_demo_ablation} compares two few-shot regimes: using 100 training images per task versus 10.
With 100 images (the same setting as in Tab.~\ref{tab:simplerenv_widowx}), world model RL boosts the overall success rate from 23.3 to 41.0, while the extreme 10-image setting still reaches 34.7 over the SFT-only baseline.
We observe improvements on grasp and full-task success across four tasks in this highly data-starved regime, suggesting that \fa{} with \rl{} can effectively leverage a small number of training images to refine the policy via world model rollouts.

\subsubsection{Evaluation on real-robot}
Tab.~\ref{tab:real_world_exp} reports real-robot results on our UR30e setup.
We fine-tune \wm{} on data for 20k steps and use it to train policies with 100 RL updates per task, initialized from 20 image snapshots.
Across the VLA variants, success increases by 24--30\% with fewer updates than SFT (100 RL steps versus 50k SFT), indicating that \fa{} and \rl{} can effectively refine policies from \wm{} rollouts in the low-data regime.
\input{tables/real_world_exp}
\FloatBarrier

\noindent\textbf{Behavior emergence in \emph{PlaceBowl}}
Fig.~\ref{fig:rl_demo_pb} illustrates a clear change in policy behavior after RL.
Although the demonstrations only contain left-side grasps, the SFT policy is stochastic and can, with very low probability, generate a right-side approach.
These rare trajectories are often unstable, but whenever a right-side attempt succeeds the RM assigns it a positive signal.
RL amplifies this weak mode in the SFT action distribution, turning the right-side strategy into a consistent and reliable behavior.
This highlights a key difference between SFT and RL: SFT imitates the dataset, whereas RL can discover and reinforce behaviors that are only weakly expressed in the demonstrations.

\noindent\textbf{Soft-object manipulation in \emph{PulloutTissue}}
Fig.~\ref{fig:rl_demo_pt} focuses on the \emph{PulloutTissue} task, where soft, deformable contact makes the problem particularly sensitive to approach pose.
The SFT policy (Pi0.5) can reproduce the overall motion pattern in the demonstrations, but is highly sensitive to small variations in the approach trajectory: even minor lateral drift in the gripper pose leads to poor contact with the tissue edge and frequent failures.
During RL, the RM assigns higher scores to trajectories that achieve clean edge engagement and complete the pull.
This provides a learning signal that reduces lateral variance and reinforces approach trajectories with better stability.
The post-RL policy exhibits a tighter distribution over approach poses and maintains alignment with the tissue edge more reliably, enabling consistent extraction across diverse initial placements.
Such fine-grained alignment is difficult to capture in standard rigid-body simulators, but \wm{}, trained on real trajectories, can represent these deformable-contact effects without hand-crafted physics.

\input{images/new_behaviour_demo}

\noindent\textbf{Closed-loop \wm{} rollouts during RL}
Fig.~\ref{fig:rl_demo_wm_rollout_tissue} visualizes closed-loop rollouts produced by \wm{} when training Pi0.5 on \emph{PulloutTissue}.
For each initial tissue-box pose, the policy interacts only with \wm{}, and we show both a successful case (S:1) and a failure case (S:0).
This diversity of predicted trajectories is crucial for GRPO-style methods: the RM can provide feedback on a wide range of behaviors, allowing the policy to reshape its action distribution rather than overfitting to a single nominal pattern.
A notable effect appears in the fourth row.
Our data are collected with a human-in-the-loop teleoperation interface, so demonstrators occasionally adjust the object position mid-trajectory.
These corrections are implicitly absorbed by \wm{} during training, and the model sometimes predicts similar object adjustments in its rollouts.
Although such behaviors are not explicitly supervised, they emerge in a purely data-driven manner and highlight the fidelity of the learned world model to nuances in the teleoperated demonstrations.

\subsubsection{Evaluation on LIBERO}
We consider two regimes:
\textbf{(i)} RL in the simulator; and
\textbf{(ii)} RL with the world model, followed by evaluation in the simulator.
All runs use training scenes to avoid leakage, and we report averages over three seeds.

\input{images/rl_tissue}
\input{images/rl_wm_rollout_tissue}
For RL in the simulator, policies are optimized for 500 RL steps.
We record the step at which the peak validation score is first reached.
As shown in Tab.~\ref{tab:rl_ablation}, our \rl{} consistently speeds up convergence and raises the final success rate across categories.
For RL with the world model, we use the \wm{} fine-tuned for 30k steps on LIBERO without action frame conditioning (not available under the online servo-control setup).
During training, the policy interacts with \wm{} for rollouts up to 500 frames.
Because long-horizon closed-loop rollouts accumulate model and reward errors and each update is slower than a simulator step, we train for 100 RL updates and select the best checkpoint within this budget.
Even though rewards come from an external RM rather than environment signals, this setting still yields clear gains over the SFT policy, and \rl{} further improves results.

\input{tables/rl_ablation_libero}
\input{tables/flowscale_sensitivity}

These gains are smaller than in the simulator, as expected: geometric and contact drift over long rollouts, combined with bias from the learned RM, make credit assignment harder in \wm{}.
Rather than replacing high-fidelity simulators, \wm{} targets regimes where such simulators are unavailable or costly.
In our experiments, it nevertheless provides training signals that are strong enough to improve policy performance under this shorter, noisier training regime.

Tab.~\ref{tab:flowscale_sens} further studies the sensitivity of \rl{} on LIBERO Spatial.
The default setting improves SuccGain from 9.7 to 11.9 and reaches peak validation performance $3.4\times$ faster than \fa{} alone.
Nearby choices of $p$, $\alpha$, and clipping remain effective, while tighter clipping reaches the peak in only 49 updates, corresponding to an $8.3\times$ speedup.

\subsection{Reward model discussion}
\label{sec:rm_discussion}

To understand what makes an RM usable for RL, we run an experiment where RL is performed in LIBERO simulator, but the policy receives rewards only from the RM.
We start from a SmolVLA~\cite{shukor2025smolvla} policy SFT on full LIBERO for 100k steps, and then post-train it on LIBERO-Spatial using \fa{}.
The simulator is used to render observations and to provide ground truth success labels for logging and RM evaluation.

\input{tables/rm_ablation}

Tab.~\ref{tab:sim_rm_control} isolates the reward source in simulator RL.
Using the ground-truth simulator reward gives the strongest result, while the 72B RM still improves substantially over SFT, confirming that RM feedback provides a usable but imperfect training signal.
Tab.~\ref{tab:rm_cost} further studies RM capacity and voting.
Larger RMs perform better in this setting, and voting improves reliability at the cost of higher latency and GPU hours.
To avoid reward-source confounds, our component ablations use the same fixed RM across variants.

\noindent\textbf{Other attempted RMs}
We also explored two RM designs that we did not adopt.
First, we generated synthetic failure rollouts by editing successful action sequences and used these rollouts as supervised negative examples for RM training.
Although the resulting videos can look plausible, as in Fig.~\ref{fig:simulated_failures}, their statistics differ from both natural successes and real-robot failures.
In our experiments, a fine-tuned VLM RM tended to exploit style cues from the edited rollouts rather than task semantics, and the training loss quickly collapsed.
We therefore do not use world-model-generated failures as RM training labels for real-world tasks, keeping the Qwen2.5-VL-72B~\cite{bai2025qwen2} RM in a zero-shot evaluation regime.
Second, we tried VLAC~\cite{zhai2025vision}, initialized from its released 2B checkpoint and fine-tuned on small amounts of real-world data.
In our setup, VLAC-2B underperformed Qwen2.5-VL-72B on subtle object interactions, so it is not used in the main experiments.

Fig.~\ref{fig:rm_diagnostics} summarizes the results.
From left to right, the columns report:
\textbf{(i)} test success rate on held-out scenes under the true simulator reward (ground truth-based success rate);
\textbf{(ii)} the fraction of on-policy training trajectories that the RM classifies as successful (RM-based success rate); and
\textbf{(iii)}--\textbf{(v)} RM precision, recall (true positive rate, TPR), and false positive rate (FPR) measured on on-policy rollouts:
\begin{equation}
\begin{aligned}
\mathrm{Precision}
  &= \Pr(\text{true success}\mid\text{RM predicts success}),\\
\mathrm{Recall}
  &= \Pr(\text{RM predicts success}\mid\text{true success}),\\
\mathrm{FPR}
  &= \Pr(\text{RM predicts success}\mid\text{true failure}).
\end{aligned}
\end{equation}

\noindent\textbf{RM training setups}
All three RMs in Fig.~\ref{fig:rm_diagnostics} share the same architecture and objective, but differ in the policies used to generate training trajectories and in the total amount of data.
We use SmolVLA checkpoints with different ground truth success rates on LIBERO-Spatial to probe how RM quality affects RL.

The low-recall RM (bottom row) is trained on 5k trajectories collected from a single checkpoint with roughly $45\%$ success.
The two high-recall RMs (top and middle rows) are trained on mixed data from two checkpoints: one with $\approx 45\%$ success and one with $\approx 70\%$ success.
For the short-run setting (middle row), we use 5k trajectories in total, 2.5k from each checkpoint.
For the long-run setting (top row), we double this to 10k trajectories, 5k per checkpoint.
In all cases, LIBERO ground truth success is used only as a label for RM training and evaluation. RL relies solely on RM-based rewards.

\noindent\textbf{High-recall RM, long run (top row)}
In the first $300$–$350$ updates, ground truth-based success improves from roughly $0.55$ to $0.7$, while RM-based success steadily increases.
Precision remains high ($\approx 0.85$–$0.9$) and recall is almost perfect ($\approx 0.98$), even though FPR drifts upward from $\sim 0.08$ to $\sim 0.3$.
In this regime, the RM is slightly over-optimistic on the shrinking set of failures, but still recognizes most truly successful trajectories and keeps most RM-labelled successes correct, so \fa{} can exploit the induced ranking to improve the policy despite moderate noise around failures.

After about $300$ updates, the dynamics change: ground truth-based success stops improving and eventually declines, RM-based success saturates, and precision starts to drop while FPR remains high.
The RM now assigns `success' to a much larger fraction of failures, and RM-labelled positives are no longer dominated by genuinely successful episodes.
Gradients from RM-based rewards become misaligned with the true task, so continued optimization on the RM signal actively hurts performance.

\input{images/reward_curve}

\noindent\textbf{High-recall RM, short run (middle row)}
The second run uses the same high-recall RM but is stopped after $200$ updates.
Over this horizon, ground truth-based success increases monotonically, RM-based success rises smoothly, precision stays above $\sim 0.85$, recall remains close to $0.98$, and FPR grows to $\sim 0.35$.
The run stays in the `useful' regime of the top-row curves and never enters the late-stage precision collapse.
This illustrates that \fa{} can tolerate increases in FPR and distributional shift, as long as the RM continues to assign high scores to most truly successful trajectories and keeps the set of RM-labelled successes reasonably clean.

\noindent\textbf{Low-recall RM, short run (bottom row)}
The third run highlights a complementary failure mode.
Here the RM has consistently low FPR ($\approx 0.1$–$0.2$) and precision comparable to the other runs ($\sim 0.8$), but its recall is significantly lower, fluctuating around $0.7$–$0.8$ instead of being near $1.0$.
Ground truth-based success now fails to improve and oscillates around the initial $0.5$ level, even though the RM appears `conservative' by rarely misclassifying failures as successes.
From a classification viewpoint, many genuinely successful trajectories are treated as negatives and receive no advantage over mediocre ones, the RM provides very weak preference between good and average behavior, and the policy has little signal to move towards true success.

Taken together, these diagnostics clarify what matters for RM quality in our setting.
Moderate increases in FPR are acceptable as the policy distribution drifts:
\fa{} can still make progress as long as the RM maintains high recall, so that most successful trajectories are recognized as such, and reasonably high precision, so that RM-labelled successes are not dominated by failures.
In contrast, two situations are harmful:
\textbf{(i)} a late-stage sharp drop in precision while FPR is high, which leads to strong misalignment between RM rewards and true success (top row, late phase); and
\textbf{(ii)} persistently low recall, even with low FPR, which yields too few correctly rewarded successes to drive learning (bottom row).
Thus, \emph{a useful RM for \fa{} should reliably find most truly successful trajectories (high recall) while keeping its precision reasonably stable. Driving FPR to very small values is neither necessary nor sufficient for good RL performance.}

%% file: tables/pretraining_results.tex
\begin{table}[t]
  \centering
  \caption{\textbf{Evaluation of the pretrained \wm{} on multiple datasets.} }
  \label{tab:pretrain}
  \small
  \scalebox{0.9}{
  \begin{tabular}{
      l
      c c c   
      c c c c 
  }
    \toprule
    \multicolumn{1}{c}{\textbf{Datasets}} &
    \multicolumn{3}{c}{\textbf{Visual fidelity} $\uparrow$} &
    \multicolumn{4}{c}{\textbf{Action consistency}} \\
    \cmidrule(lr){2-4} \cmidrule(lr){5-8}
    & \multicolumn{1}{c}{$\mathrm{PSNR}$} &
      \multicolumn{1}{c}{$\mathrm{SSIM}$} &
      \multicolumn{1}{c}{$\mathrm{tSSIM}$} &
      \multicolumn{1}{c}{$\overline{\mathrm{EPE}}\,\downarrow$} &
      \multicolumn{1}{c}{$\widetilde{\mathrm{EPE}}\,\downarrow$} &
      \multicolumn{1}{c}{$\overline{\cos}\,\uparrow$} &
      \multicolumn{1}{c}{$\widetilde{\cos}\,\uparrow$} \\
    \midrule
      AgiBot~\cite{bu2025AgiBot}  & 27.05 & .8916 & .7666 & .2959 & .2750 & .2144 & .2176 \\
      DROID~\cite{khazatsky2024droid} & 25.23 & .8813 & .7812 & .2574 & .2385 & .1532 & .1568 \\
      Open-X~\cite{o2024open}     & 27.25 & .8810 & .7950 & .4521 & .2910 & .0843 & .0822 \\
      LIBERO~\cite{liu2023libero} & 26.29 & .9075 & .8639 & .1660 & .1602 & .4164 & .4235 \\
    \bottomrule
  \end{tabular}
  }
\end{table}

%% file: tables/bridge_finetuning.tex
\begin{table}[!htbp]
  \centering
  \caption{\textbf{World model evaluation on BRIDGE.} The first three tasks evaluate \emph{same-task} few-shot fine-tuning with evaluation on larger held-out splits; the middle three evaluate \emph{cross-object} transfer, where few-shot fine-tuning is performed on object instance of the same task and evaluation is on unseen objects of that task; the last one evaluates the full-data fine-tuning results. \textbf{Bold} and \underline{underlined} mark the best and second-best (same applies to the following tables), and `train/val' denotes the number of trajectories used for fine-tuning / evaluation.}
  \label{tab:bridge_finetune}
  \small
  \scalebox{0.85}{
  \begin{tabular}{
      l l l
      S[table-format=2.3] 
      S[table-format=1.3] 
      S[table-format=1.3] 
      S[table-format=2.3] 
      S[table-format=2.3] 
      S[table-format=1.3] 
      S[table-format=1.3] 
  }
    \toprule
    \multicolumn{1}{c}{\textbf{Method}} &
    \multicolumn{1}{c}{\textbf{Train-val}} &
    \multicolumn{1}{c}{\textbf{Task(s)}} &
    \multicolumn{3}{c}{\textbf{Visual fidelity} $\uparrow$} &
    \multicolumn{4}{c}{\textbf{Action consistency}} \\
    \cmidrule(lr){4-6} \cmidrule(lr){7-10}
    & & &
    \multicolumn{1}{c}{$\mathrm{PSNR}$} &
    \multicolumn{1}{c}{$\mathrm{SSIM}$} &
    \multicolumn{1}{c}{$\mathrm{tSSIM}$} &
    \multicolumn{1}{c}{$\overline{\mathrm{EPE}}$ $\downarrow$} &
    \multicolumn{1}{c}{$\widetilde{\mathrm{EPE}}$ $\downarrow$} &
    \multicolumn{1}{c}{$\overline{\mathrm{cos}}$ $\uparrow$} &
    \multicolumn{1}{c}{$\widetilde{\mathrm{cos}}$ $\uparrow$} \\
    \midrule
    
    \multicolumn{10}{l}{\textit{Experiment 1: same-task few-shot transfer}} \\
    \midrule
    \multirow{2}{*}{LTX-Video~\cite{HaCohen2024LTXVideo}}
        & 100-2737 & \texttt{fold cloths}     & \underline{21.44} & {.7549} & {.5098} & {1.8329} & \underline{1.4782} & {.0397} & {.0340} \\
      & 150-331 & \texttt{close the drawer}     & \underline{22.32} & {.7565} & {.5052} & {1.4707} & \underline{1.2827} & {.0673} & {.0617} \\
      & 150-856 & \texttt{remove lid} & {22.27} & {.7429} & {.5082} & \underline{1.2379} & \underline{0.9502} & {.0295} & {.0281} \\
      
    \arrayrulecolor{gray!60}\midrule\arrayrulecolor{black}
    \multirow{2}{*}{Genie-envisioner~\cite{liao2025genie}}
      & 100-2737 & \texttt{fold cloths} & \textbf{21.96} & \underline{.7918} & {.6009} & \underline{1.8306} & {1.5534} & \underline{.0813} & \underline{.0873} \\
      & 150-331 & \texttt{close the drawer}     & \textbf{22.93} & \underline{.7898} & {.5764} & {1.5163} & {1.3813} & {.1172} & {.1130} \\
      & 150-856 & \texttt{remove lid}     & \textbf{23.60} & \underline{.7924} & {.5705} & {1.2390} & {1.0012} & {.0661} & {.0690} \\
      
    \arrayrulecolor{gray!60}\midrule\arrayrulecolor{black}
    \multirow{2}{*}{Cosmos-Predict2~\cite{agarwal2025cosmos}}
      & 100-2737 & \texttt{fold cloths}     & {19.44} & {.7723} & \underline{.6205} & {1.9468} & {1.5755} & {.0745} & {.0766} \\
      & 150-331 & \texttt{close the drawer}     & {19.91} & {.7737} & \underline{.6171} & \underline{1.4515} & {1.3058} & \underline{.1389} & \underline{.1455} \\
      & 150-856 & \texttt{remove lid}     & {21.70} & {.7798} & \underline{.6429} & {1.3404} & {1.0601} & \underline{.0711} & \underline{.0756} \\
      
    \arrayrulecolor{gray!60}\midrule\arrayrulecolor{black}
    \multirow{2}{*}{\wm{}~\textit{(Ours)}}
      & 100-2737 & \texttt{fold cloths}     & {21.30} & \textbf{.7945} & \textbf{.6481} & \textbf{1.4489} & \textbf{1.1768} & \textbf{.1836} & \textbf{.1870} \\
      & 150-331 & \texttt{close the drawer}     & {21.96} & \textbf{.7999} & \textbf{.6505} & \textbf{1.0890} & \textbf{0.9508} & \textbf{.2225} & \textbf{.2266} \\
      & 150-856 & \texttt{remove lid} & \underline{23.58} & \textbf{.8036} & \textbf{.6795} & \textbf{0.9407} & \textbf{0.6990} & \textbf{.1312} & \textbf{.1342} \\
      
    \midrule
    \multicolumn{10}{l}{\textit{Experiment 2: cross-object few-shot transfer}} \\
    \midrule
    \multirow{2}{*}{LTX-Video~\cite{HaCohen2024LTXVideo}}
      & 150-33 & \texttt{move sth.} $\Rightarrow$ \texttt{move carrot} & {21.97} & {.7335} & {.4706} & {1.5567} & {1.3250} & {.0238} & {.0226} \\
      & 150-50 & \texttt{put cube} $\Rightarrow$ \texttt{put yellow cube} & {21.11} & {.6684} & {.4669} & {1.3699} & {1.0701} & {.0415} & {.0435} \\
      & 150-59 & \texttt{pick up sth.} $\Rightarrow$ \texttt{pick up spoon} & {21.37} & {.7250} & {.4664} & {1.4971} & {1.2296} & {.0257} & {.0280} \\

    \arrayrulecolor{gray!60}\midrule\arrayrulecolor{black}
    \multirow{2}{*}{Genie-envisioner~\cite{liao2025genie}}
      & 150-33 & \texttt{move sth.} $\Rightarrow$ \texttt{move carrot} & \textbf{22.32} & \textbf{.7769} & {.5407} & {1.6545} & {1.4396} & {.0424} & {.0406} \\
      & 150-50 & \texttt{put cube} $\Rightarrow$ \texttt{put yellow cube} & \underline{22.15} & \underline{.7538} & {.5551} & {1.4554} & {1.1870} & \underline{.0795} & \underline{.0815} \\
      & 150-59 & \texttt{pick up sth.} $\Rightarrow$ \texttt{pick up spoon} & \underline{22.05} & \underline{.7700} & {.5324} & {1.5353} & {1.3067} & \underline{.0610} & \underline{.0657} \\
      
    \arrayrulecolor{gray!60}\midrule\arrayrulecolor{black}
    \multirow{2}{*}{Cosmos-Predict2~\cite{agarwal2025cosmos}}
      & 150-33 & \texttt{move sth.} $\Rightarrow$ \texttt{move carrot} & {19.80} & {.7476} & \underline{.5827} & {1.8921} & {1.5433} & \underline{.0479} & \underline{.0489} \\
      & 150-50 & \texttt{put cube} $\Rightarrow$ \texttt{put yellow cube} & {19.89} & {.7398} & \underline{.6138} & {1.7488} & {1.2797} & {.0640} & {.0664} \\
      & 150-59 & \texttt{pick up sth.} $\Rightarrow$ \texttt{pick up spoon} & {19.18} & {.7582} & \underline{.6090} & {1.9190} & {1.5245} & {.0576} & {.0583} \\
      
    \arrayrulecolor{gray!60}\midrule\arrayrulecolor{black}
    \multirow{2}{*}{\wm{}~\textit{(Ours)}}
      & 150-33 & \texttt{move sth.} $\Rightarrow$ \texttt{move carrot} & \underline{22.09} & \underline{.7739} & \textbf{.6128} & \textbf{1.4466} & \textbf{1.1315} & \textbf{.1312} & \textbf{.1310} \\
      & 150-50 & \texttt{put cube} $\Rightarrow$ \texttt{put yellow cube} & \textbf{23.15} & \textbf{.7887} & \textbf{.6666} & \textbf{1.0311} & \textbf{0.7216} & \textbf{.1750} & \textbf{.1782} \\
      & 150-59 & \texttt{pick up sth.} $\Rightarrow$ \texttt{pick up spoon} & \textbf{22.34} & \textbf{.7944} & \textbf{.6535} & \textbf{1.1945} & \textbf{0.9254} & \textbf{.1430} & \textbf{.1473} \\

    \midrule
    \multicolumn{10}{l}{\textit{Experiment 3: Full-data fine-tuning}} \\
    \midrule
    Ctrl-World~\cite{guo2025ctrl}
      & 36243-2417 & \texttt{full-data}     & {21.87} & {.8172} & \textbf{.7261} & {1.0674} & \underline{0.7721} & {.1419} & {.1327} \\
    SVD~\cite{blattmann2023stable}
      & 36243-2417 & \texttt{full-data}     & {20.95} & {.8032} & \underline{.7157} & {1.2587} & {0.7843} & {.1204} & {.1113} \\
    LTX-Video~\cite{HaCohen2024LTXVideo}
      & 36243-2417 & \texttt{full-data}     & {23.34} & {.7970} & {.5739} & {1.4289} & {1.1657} & {.0980} & {.1005} \\
    Genie-envisioner~\cite{liao2025genie}
      & 36243-2417 & \texttt{full-data}     & {23.66} & {.8038} & {.5825} & {1.4328} & {1.1950} & {.1061} & {.1107} \\
    IRASim~\cite{zhu2024irasim}
      & 36243-2417 & \texttt{full-data}     & \underline{25.00} & {.8146} & {.6510} & \underline{0.9558} & {0.8188} & {.1765} & {.1788} \\
    Cosmos-Predict2~\cite{agarwal2025cosmos}
      & 36243-2417 & \texttt{full-data}     & {24.58} & \underline{.8276} & {.6723} & {1.0243} & {0.8121} & \underline{.2051}  & \underline{.2071} \\
    \wm{}~\textit{(Ours)}
      & 36243-2417 & \texttt{full-data}     & \textbf{25.47} & \textbf{.8367} & {.6839} & \textbf{0.9136} & \textbf{0.7378} & \textbf{.2234} & \textbf{.2245} \\
      
    \bottomrule
  \end{tabular}
  }
  \footnotesize

\end{table}

%% file: tables/sii_finetuning.tex
\begin{table}[!htbp]
  \centering
  \caption{\textbf{Evaluation of the fine-tuned \wm~on custom dataset.} `\textbf{Hist.}' for history-aware mechanism, `\textbf{Pre.}' denotes pretraining, and `\textbf{A-frm.}' for adding action-frame conditioning (same applies to the following tables).}
  \label{tab:sii_finetune}
  \small
  \scalebox{.9}{
  \begin{tabular}{
      c c c            
      c 
      c 
      c 
      c 
      c 
      c 
      c 
  }
    \toprule
    \multicolumn{3}{c}{\textbf{Variants}} &
    \multicolumn{3}{c}{\textbf{Visual fidelity} $\uparrow$} &
    \multicolumn{4}{c}{\textbf{Action consistency}} \\
    \cmidrule(lr){1-3}\cmidrule(lr){4-6}\cmidrule(lr){7-10}
    \textbf{Hist.} & \textbf{Pre.} & \textbf{A-frm.} &
    \multicolumn{1}{c}{$\mathrm{PSNR}$} &
    \multicolumn{1}{c}{$\mathrm{SSIM}$} &
    \multicolumn{1}{c}{$\mathrm{tSSIM}$} &
    \multicolumn{1}{c}{$\overline{\mathrm{EPE}}\,\downarrow$} &
    \multicolumn{1}{c}{$\widetilde{\mathrm{EPE}}\,\downarrow$} &
    \multicolumn{1}{c}{$\overline{\cos}\,\uparrow$} &
    \multicolumn{1}{c}{$\widetilde{\cos}\,\uparrow$} \\
    \midrule
     \xmark & \xmark & \xmark   & 24.28 & .8952 & .7807 & .5396 & .4907 & .1830 & .1865 \\
     \cmark & \xmark & \xmark   & 24.85 & .9008 & .7881 & .5224 & .4711 & .1959 & .1996 \\
     \cmark & \cmark & \xmark   & \underline{25.92} & \underline{.9123} & \underline{.7995} & \underline{.4463} & \underline{.4061} & \underline{.2157} & \underline{.2191} \\
     \cmark & \cmark & \cmark   & \textbf{26.12} & \textbf{.9150} & \textbf{.8032} & \textbf{.4345} & \textbf{.3952} & \textbf{.2189} & \textbf{.2214} \\
    \bottomrule
  \end{tabular}
  }
\end{table}

%% file: tables/libero_finetuning.tex
\begin{table}[!htbp]
  \centering
  \caption{\textbf{Ablation of \wm~components on LIBERO.} Tasks: \texttt{S} = Spatial, \texttt{O} = Object, \texttt{G} = Goal, \texttt{L} = Long.}
  \label{tab:libero_ablation}
  \small
  \scalebox{.75}{
  \begin{tabular}{
      c c c            
      c                
      c 
      c 
      c 
      c 
      c 
      c 
      c 
  }
    \toprule
    \multicolumn{3}{c}{\textbf{Variants}} &
    \multicolumn{1}{c}{\textbf{Task}} &
    \multicolumn{3}{c}{\textbf{Visual fidelity} $\uparrow$} &
    \multicolumn{4}{c}{\textbf{Action consistency}} \\
    \cmidrule(lr){1-3}\cmidrule(lr){5-7}\cmidrule(lr){8-11}
    \textbf{Hist.} & \textbf{Pre.} & \textbf{A-frm.} &
    &
    \multicolumn{1}{c}{$\mathrm{PSNR}$} &
    \multicolumn{1}{c}{$\mathrm{SSIM}$} &
    \multicolumn{1}{c}{$\mathrm{tSSIM}$} &
    \multicolumn{1}{c}{$\overline{\mathrm{EPE}}\,\downarrow$} &
    \multicolumn{1}{c}{$\widetilde{\mathrm{EPE}}\,\downarrow$} &
    \multicolumn{1}{c}{$\overline{\cos}\,\uparrow$} &
    \multicolumn{1}{c}{$\widetilde{\cos}\,\uparrow$} \\
    \midrule

    \xmark & \xmark & \xmark & \texttt{S} & {34.29} & {.9683} & {.9221} & {.0771} & {.0744} & {.5332} & {.5366} \\
    \cmark & \xmark & \xmark & \texttt{S} & {34.44} & {.9690} & {.9239} & {.0752} & {.0727} & {.5359} & {.5399} \\
    \cmark & \cmark & \xmark & \texttt{S} & \textbf{34.66} & \textbf{.9699} & \underline{.9239} & \underline{.0740} & \underline{.0714} & \underline{.5393} & \underline{.5429} \\
    \cmark & \cmark & \cmark & \texttt{S} & \underline{34.54} & \underline{.9697} & \textbf{.9256} & \textbf{.0730} & \textbf{.0706} & \textbf{.5401} & \textbf{.5437} \\
    \addlinespace[4pt]
    \arrayrulecolor{gray!60}\midrule\arrayrulecolor{black}

    \xmark & \xmark & \xmark & \texttt{O}  & {32.91} & {.9556} & {.9168} & {.0681} & {.0630} & {.3338} & {.3350} \\
    \cmark & \xmark & \xmark & \texttt{O}  & {33.01} & {.9565} & \underline{.9192} & \underline{.0665} & \underline{.0617} & {.3378} & \underline{.3396} \\
    \cmark & \cmark & \xmark & \texttt{O}  & \underline{33.10} & \underline{.9567} & .9177 & {.0677} & {.0619} & \underline{.3381} & \underline{.3396} \\
    \cmark & \cmark & \cmark & \texttt{O}  & \textbf{33.15} & \textbf{.9575} & \textbf{.9210} & \textbf{.0649} & \textbf{.0603} & \textbf{.3416} & \textbf{.3432} \\
    
    \addlinespace[4pt]
    \arrayrulecolor{gray!60}\midrule\arrayrulecolor{black}
    \xmark & \xmark & \xmark & \texttt{G}    & {35.14} & {.9704} & {.9329} & {.0661} & {.0633} & {.4447} & {.4496} \\
    \cmark & \xmark & \xmark & \texttt{G}    & {35.22} & {.9707} & \underline{.9345} & {.0649} & {.0622} & {.4465} & {.4508} \\
    \cmark & \cmark & \xmark & \texttt{G}    & \textbf{35.44} & \textbf{.9717} & .9339 & \underline{.0623} & \underline{.0593} & \underline{.4519} & \underline{.4558} \\
    \cmark & \cmark & \cmark & \texttt{G}    & \underline{35.42} & \textbf{.9717} & \textbf{.9362} & \textbf{.0612} & \textbf{.0583} & \textbf{.4528} & \textbf{.4562} \\
    \addlinespace[4pt]
    \arrayrulecolor{gray!60}\midrule\arrayrulecolor{black}

    \xmark & \xmark & \xmark & \texttt{L}    & {32.53} & {.9597} & {.9124} & {.1015} & {.0980} & {.4253} & {.4275} \\
    \cmark & \xmark & \xmark & \texttt{L}    & {32.86} & {.9620} & {.9160} & {.0976} & {.0944} & {.4289} & {.4314} \\
    \cmark & \cmark & \xmark & \texttt{L}    & \textbf{33.18} & \underline{.9631} & \underline{.9164} & \underline{.0929} & \underline{.0895} & \textbf{.4348} & \textbf{.4374} \\
    \cmark & \cmark & \cmark & \texttt{L}    & \underline{33.09} & \textbf{.9635} & \textbf{.9185} & \textbf{.0924} & \textbf{.0894} & \underline{.4347} & \underline{.4372} \\
    \bottomrule
  \end{tabular}
}
\end{table}

%% file: tables/simplerenv.tex
\begin{table}[!htbp]
\centering
\caption{\textbf{SimplerEnv (WidowX) evaluation with RL post-training on BRIDGE.}
For reproduced VLA variants, policies take a single RGB image (no multi-view, history, or state) as input, and SFT for 200k steps. 
Rollouts from our \wm{} are used for decision support.}
\resizebox{1\textwidth}{!}{
\begin{tabular}{l|ccccccccccc}
\toprule
\multirow{2}{*}{\textbf{Method}} &
\multicolumn{2}{c}{\textbf{Put Spoon on Towel}} &
\multicolumn{2}{c}{\textbf{Put Carrot on Plate}} &
\multicolumn{2}{c}{\textbf{Stack Green Block on Yellow Block}} &
\multicolumn{2}{c}{\textbf{Put Eggplant in Yellow Basket}} &
\textbf{Partial} &
\textbf{Overall} \\
& Grasp Spoon & Success & Grasp Carrot & Success & Grasp Green Block & Success & Grasp Eggplant & Success & Average & Average \\
\cmidrule{1-11}
RT-1-X~\cite{o2024open} & 16.7 & 0 & 20.8 & 4.2 & 8.3 & 0 & 0.0 & 0 & 11.5 & 1.1 \\
Octo-Base~\cite{octo2023} & 34.7 & 12.5 & 52.8 & 8.3 & 31.9 & 0 & 66.7 & 43.1 & 46.5 & 16.0 \\
Octo-Small~\cite{octo2023} & 77.8 & 47.2 & 27.8 & 9.7 & 40.3 & 4.2 & 87.5 & 56.9 & 58.4 & 30.0 \\
OpenVLA~\cite{kim2025openvla} & 4.1 & 0 & 33.3 & 0 & 12.5 & 0 & 8.3 & 4.1 & 14.6 & 1.0 \\
RoboVLM (fine-tuning)~\cite{liu2025towards} & 54.2 & 29.2 & 25.0 & 25.0 & 45.8 & 12.5 & 58.3 & 58.3 & 45.8 & 31.3 \\
SpatialVLA (fine-tuning)~\cite{qu2025spatialvla} & 20.8 & 16.7 & 29.2 & 25.0 & 62.5 & 29.2 & 100.0 & 100.0 & 53.1 & 42.7 \\
\midrule
VLA-Adapter-0.5B~\cite{wang2025vla} &
$45.9\pm 7.2$ & $18.0\pm 8.7$ & $40.3\pm 6.4$ & $18.1\pm 4.8$ &
$69.4\pm 6.4$ & $7.0\pm 4.8$ & $72.2\pm 7.2$ & $50.0\pm 4.8$ &
$57.0\pm 5.1$ & $23.3\pm 2.2$ \\

\hspace{0.9em}+\,\fa \textit{~(Ours)} &
$\underline{66.7}\pm 7.2\,\deltaup{+20.8}$ &
$\mathbf{38.9}\pm 4.8\,\deltaup{+20.9}$ &
$\underline{45.8}\pm 6.4\,\deltaup{+5.5}$ &
$\underline{29.2}\pm 4.2\,\deltaup{+11.1}$ &
$\underline{76.4}\pm 8.4\,\deltaup{+7.0}$ &
$\underline{9.7}\pm 2.4\,\deltaup{+2.7}$ &
$\mathbf{88.9}\pm 8.4\,\deltaup{+16.7}$ &
$\underline{75.0}\pm 8.4\,\deltaup{+25.0}$ &
$\underline{69.4}\pm 4.8\,\deltaup{+12.4}$ &
$\underline{38.2}\pm 2.4\,\deltaup{+14.9}$ \\

\hspace{0.9em}+\,\fa{}~\&~\rl{}\textit{~(Ours)} &
$\mathbf{70.8}\pm 7.2\,\deltaup{+24.9}$ &
$\underline{33.3}\pm 4.2\,\deltaup{+15.3}$ &
$\mathbf{52.8}\pm 2.4\,\deltaup{+12.5}$ &
$\mathbf{36.1}\pm 4.8\,\deltaup{+18.0}$ &
$\mathbf{77.8}\pm 4.8\,\deltaup{+8.4}$ &
$\mathbf{15.3}\pm 4.8\,\deltaup{+8.3}$ &
$\underline{87.5}\pm 8.3\,\deltaup{+15.3}$ &
$\mathbf{79.2}\pm 4.8\,\deltaup{+29.2}$ &
$\mathbf{72.2}\pm 2.4\,\deltaup{+15.2}$ &
$\mathbf{41.0}\pm 2.4\,\deltaup{+17.7}$ \\

\midrule
Pi0.5-3B~\cite{black2025pi_} &
$65.3\pm 6.4$ & $44.4\pm 6.4$ & $\underline{57.0}\pm 4.8$ & $29.2\pm 0$ &
$75.0\pm 7.2$ & $\underline{18.1}\pm 6.4$ & $80.5\pm 4.8$ & $63.9\pm 2.4$ &
$69.5\pm 2.4$ & $38.9\pm 2.6$ \\

\hspace{0.9em}+\,\fa \textit{~(Ours)} &
$\mathbf{75.8}\pm 3.0\,\deltaup{+10.5}$ &
$\underline{51.4}\pm 4.8\,\deltaup{+7.0}$ &
$\mathbf{59.7}\pm 8.7\,\deltaup{+2.7}$ &
$\underline{41.6}\pm 7.2\,\deltaup{+12.4}$ &
$\mathbf{91.7}\pm 4.2\,\deltaup{+16.7}$ &
$\mathbf{22.2}\pm 6.4\,\deltaup{+4.1}$ &
$\underline{82.0}\pm 4.8\,\deltaup{+1.5}$ &
$\underline{72.2}\pm 6.4\,\deltaup{+8.3}$ &
$\mathbf{77.3}\pm 4.8\,\deltaup{+7.8}$ &
$\underline{46.9}\pm 3.0\,\deltaup{+8.0}$ \\

\hspace{0.9em}+\,\fa{}~\&~\rl{}\textit{~(Ours)} &
$\underline{72.8}\pm 5.2\,\deltaup{+7.5}$ &
$\textbf{58.3}\pm 4.8\,\deltaup{+13.9}$ &
$\textbf{59.7}\pm 4.8\,\deltaup{+2.7}$ &
$\textbf{43.0}\pm 8.7\,\deltaup{+13.8}$ &
$\underline{82.0}\pm 4.8\,\deltaup{+7.0}$ &
$\textbf{22.2}\pm 4.8\,\deltaup{+4.1}$ &
$\textbf{93.2}\pm 2.2\,\deltaup{+12.7}$ &
$\textbf{80.6}\pm 2.4\,\deltaup{+16.7}$ &
$\underline{76.9}\pm 2.8\,\deltaup{+7.4}$ &
$\textbf{51.0}\pm 1.2\,\deltaup{+12.1}$ \\

\midrule
OpenVLA-OFT-7B~\cite{kim2025openvla} &
$45.8\pm 4.2$ & $25.0\pm 4.2$ & $40.3\pm 4.8$ & $13.9\pm 2.4$ &
$50.0\pm 8.3$ & $5.6\pm 2.4$ & $79.2\pm 4.2$ & $55.5\pm 2.3$ &
$53.8\pm 3.2$ & $25.0\pm 1.8$ \\

\hspace{0.9em}+\,\fa \textit{~(Ours)} &
$\underline{46.0}\pm 7.0\,\deltaup{+0.2}$ &
$\underline{26.4}\pm 2.4\,\deltaup{+1.4}$ &
$\underline{41.7}\pm 7.2\,\deltaup{+1.4}$ &
$\mathbf{19.5}\pm 4.8\,\deltaup{+5.6}$ &
$\underline{61.1}\pm 8.6\,\deltaup{+11.1}$ &
$\underline{9.7}\pm 4.8\,\deltaup{+4.1}$ &
$\mathbf{93.1}\pm 2.4\,\deltaup{+14.3}$ &
$\underline{61.1}\pm 2.4\,\deltaup{+5.6}$ &
$\underline{60.5}\pm 2.7\,\deltaup{+6.7}$ &
$\underline{29.2}\pm 1.8\,\deltaup{+4.2}$ \\

\hspace{0.9em}+\,\fa{}~\&~\rl{}\textit{~(Ours)} &
$\mathbf{55.6}\pm 7.2\,\deltaup{+9.8}$ &
$\mathbf{29.2}\pm 4.2\,\deltaup{+4.2}$ &
$\mathbf{57.0}\pm 8.7\,\deltaup{+16.7}$ &
$\underline{16.7}\pm 4.2\,\deltaup{+2.8}$ &
$\mathbf{69.4}\pm 6.4\,\deltaup{+19.4}$ &
$\mathbf{11.1}\pm 2.4\,\deltaup{+5.5}$ &
$\underline{90.3}\pm 4.8\,\deltaup{+11.1}$ &
$\mathbf{66.7}\pm 4.2\,\deltaup{+11.2}$ &
$\mathbf{68.1}\pm 2.1\,\deltaup{+14.3}$ &
$\mathbf{30.9}\pm 0.6\,\deltaup{+5.9}$ \\
\bottomrule
\end{tabular}
}
\label{tab:simplerenv_widowx}
\end{table}

%% file: tables/rl_baseline_comparison.tex
\begin{table}[!htbp]
\centering
\begin{minipage}[t]{0.49\linewidth}
  \centering
  \captionof{table}{\textbf{RL baseline comparison.}
  Success-rate gains from the same Pi0.5 SFT policy.}
  \label{tab:rl_baseline_comparison}
  \vspace{0.2em}
  \includegraphics[width=\linewidth]{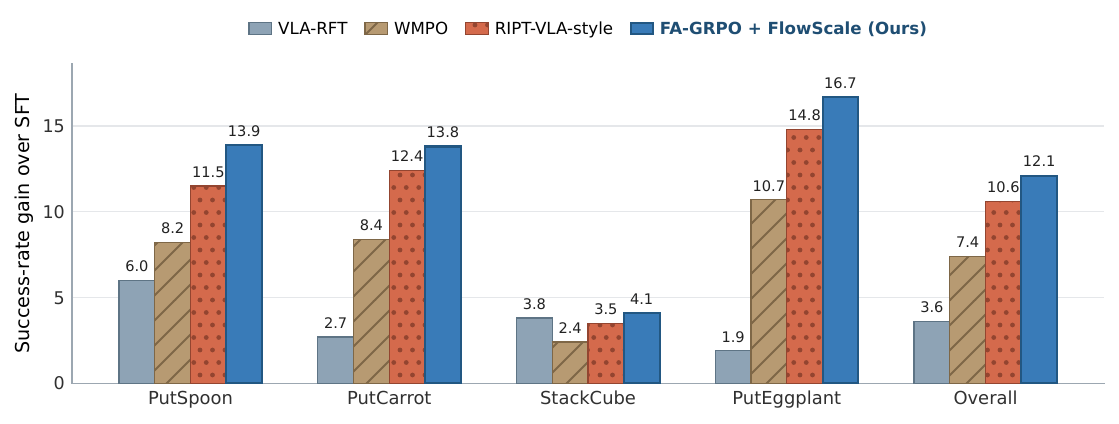}
\end{minipage}
\hfill
\addtocounter{table}{3}
\begin{minipage}[t]{0.49\linewidth}
  \centering
  \captionof{table}{\textbf{Iterative RL on PutCarrot.}
  Later rounds focus on low-reward scenes.}
  \label{tab:iterative_rl_carrot}
  \vspace{0.2em}
  \includegraphics[width=\linewidth]{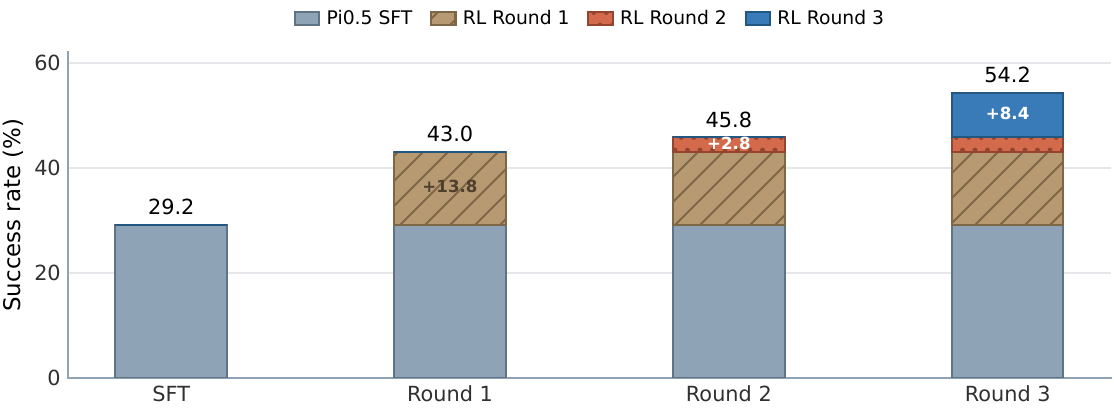}
\end{minipage}
\addtocounter{table}{-4}
\end{table}

%% file: tables/simplerenv_windowx_joint_ablation.tex
\begin{table}[!htbp]
\centering
\caption{\textbf{Multi-task RL in the world model.}
Additional results where we jointly fine-tune VLA policies (all SFT 200k steps) on the four SimplerEnv-WidowX tasks using \fa{} and \rl{} in \wm{} for 250 RL updates.
}
\resizebox{1\textwidth}{!}{
\begin{tabular}{l|cccccccccc}
\toprule
\multirow{2}{*}{Method} &
\multicolumn{2}{c}{\textbf{Put Spoon on Towel}} &
\multicolumn{2}{c}{\textbf{Put Carrot on Plate}} &
\multicolumn{2}{c}{\textbf{Stack Green Block on Yellow Block}} &
\multicolumn{2}{c}{\textbf{Put Eggplant in Yellow Basket}} &
\textbf{Partial} &
\textbf{Overall} \\
& Grasp Spoon & Success & Grasp Carrot & Success & Grasp Green Block & Success & Grasp Eggplant & Success & Average & Average \\
\midrule
VLA-Adapter-0.5B~\cite{wang2025vla} &
$45.9\pm 7.2$ & $18.0\pm 8.4$ & $40.3\pm 6.3$ & $18.1\pm 4.8$ &
$69.4\pm 6.3$ & $7.0\pm 4.8$ & $72.2\pm 7.2$ & $50.0\pm 4.8$ &
$57.0\pm 5.1$ & $23.3\pm 2.2$ \\
\hspace{0.9em}+\,\fa{}~\&~\rl{}-\textit{joint} &
$\mathbf{59.7}\pm 9.6\,\deltaup{+13.8}$ &
$\mathbf{36.1}\pm 8.6\,\deltaup{+18.1}$ &
$\mathbf{58.3}\pm 4.2\,\deltaup{+18.0}$ &
$\mathbf{29.2}\pm 4.2\,\deltaup{+11.1}$ &
$\mathbf{73.6}\pm 4.9\,\deltaup{+4.2}$ &
$\mathbf{12.5}\pm 4.2\,\deltaup{+5.5}$ &
$\mathbf{77.8}\pm 2.4\,\deltaup{+5.6}$ &
$\mathbf{65.3}\pm 4.8\,\deltaup{+15.3}$ &
$\mathbf{67.4}\pm 3.2\,\deltaup{+10.4}$ &
$\mathbf{35.8}\pm 1.6\,\deltaup{+12.5}$ \\
\midrule
Pi0.5-3B~\cite{black2025pi_} &
$65.3\pm 6.4$ & $44.4\pm 6.4$ & $57.0\pm 4.8$ & $29.2\pm 0$ &
$75.0\pm 7.2$ & $18.1\pm 6.4$ & $80.5\pm 4.8$ & $63.9\pm 2.4$ &
$69.5\pm 2.4$ & $38.9\pm 2.6$ \\

\hspace{0.9em}+\,\fa{}~\&~\rl{}-\textit{joint} &
$\mathbf{75.8}\pm 3.0\,\deltaup{+10.5}$ &
$\mathbf{51.4}\pm 4.8\,\deltaup{+7.0}$ &
$\mathbf{59.7}\pm 8.7\,\deltaup{+2.7}$ &
$\mathbf{41.6}\pm 7.2\,\deltaup{+12.4}$ &
$\mathbf{91.7}\pm 4.2\,\deltaup{+16.7}$ &
$\mathbf{22.2}\pm 6.4\,\deltaup{+4.1}$ &
$\mathbf{82.0}\pm 4.8\,\deltaup{+1.5}$ &
$\mathbf{72.2}\pm 6.4\,\deltaup{+8.3}$ &
$\mathbf{77.3}\pm 3.0\,\deltaup{+7.8}$ &
$\mathbf{46.9}\pm 4.8\,\deltaup{+8.0}$ \\

\midrule
OpenVLA-OFT-7B~\cite{kim2025openvla} &
$45.8\pm 4.2$ & $25.0\pm 4.2$ & $40.3\pm 4.8$ & $13.9\pm 2.4$ &
$50.0\pm 8.3$ & $5.6\pm 2.4$ & $79.2\pm 4.2$ & $55.5\pm 2.3$ &
$53.8\pm 3.2$ & $25.0\pm 1.8$ \\

\hspace{0.9em}+\,\fa{}~\&~\rl{}-\textit{joint} &
$\mathbf{59.7}\pm 9.6\,\deltaup{+13.9}$ &
$\mathbf{30.6}\pm 2.4\,\deltaup{+5.6}$ &
$\mathbf{59.7}\pm 2.4\,\deltaup{+19.4}$ &
$\mathbf{30.6}\pm 2.4\,\deltaup{+16.7}$ &
$\mathbf{65.3}\pm 9.6\,\deltaup{+15.3}$ &
$\mathbf{9.7}\pm 2.4\,\deltaup{+4.1}$ &
$\mathbf{83.3}\pm 11.0\,\deltaup{+4.1}$ &
$\mathbf{70.8}\pm 8.4\,\deltaup{+15.3}$ &
$\mathbf{67.0}\pm 3.3\,\deltaup{+13.2}$ &
$\mathbf{35.4}\pm 1.1\,\deltaup{+10.4}$ \\

\bottomrule
\end{tabular}
}
\label{tab:simplerenv_windowx_joint_ablation}
\end{table}

%% file: tables/simplerenv_wm_ablation.tex
\begin{table}[!htbp]
\centering
\caption{\textbf{Ablation of world model on BRIDGE.}
We compare RL trained with Cosmos-Predict2 vs.\ \wm{}, after fine-tuning each model on either the full set (30k steps) or a few-shot set (400 samples, 500 steps, denoted \textit{Few}).}

\resizebox{1\textwidth}{!}{
\begin{tabular}{l|cccccccccc}
\toprule
\multirow{2}{*}{\textbf{Method}} &
\multicolumn{2}{c}{\textbf{Put Spoon on Towel}} &
\multicolumn{2}{c}{\textbf{Put Carrot on Plate}} &
\multicolumn{2}{c}{\textbf{Stack Green Block on Yellow Block}} &
\multicolumn{2}{c}{\textbf{Put Eggplant in Yellow Basket}} &
\textbf{Partial} &
\textbf{Overall} \\
& Grasp Spoon & Success & Grasp Carrot & Success & Grasp Green Block & Success & Grasp Eggplant & Success & Average & Average \\
\midrule
VLA-Adapter-0.5B~\cite{wang2025vla} (SFT 200k steps) 
& $45.9\pm 7.2$ & $18.0\pm 8.4$ & $40.3\pm 6.3$ & $18.1\pm 4.8$ &
$69.4\pm 6.3$ & $7.0\pm 4.8$ & $72.2\pm 7.2$ & $50.0\pm 4.8$ &
$57.0\pm 5.1$ & $23.3\pm 2.2$ \\

\midrule
Cosmos-Predict2~\cite{agarwal2025cosmos}~+~\fa{}~\&~\rl{} &
$52.8\pm 6.4\,\deltaup{+6.9}$ &
$26.4\pm 2.4\,\deltaup{+8.4}$ &
$\textbf{52.8}\pm 6.4\,\deltaup{+12.5}$ &
$34.7\pm 2.4\,\deltaup{+16.6}$ &
$72.2\pm 6.4\,\deltaup{+2.8}$ &
$11.1\pm 2.4\,\deltaup{+4.1}$ &
$86.5\pm 7.2\,\deltaup{+14.3}$ &
$76.4\pm 6.4\,\deltaup{+16.4}$ &
$66.3\pm 3.7\,\deltaup{+9.3}$ &
$37.1\pm 2.2\,\deltaup{+13.8}$ \\
\wm~+~\fa{}~\&~\rl{} 
&
$\textbf{70.8}\pm 7.2\,\deltaup{+24.9}$ &
$\textbf{33.3}\pm 4.2\,\deltaup{+15.3}$ &
$\textbf{52.8}\pm 2.4\,\deltaup{+12.5}$ &
$\textbf{36.1}\pm 4.8\,\deltaup{+18.0}$ &
$\textbf{77.8}\pm 4.8\,\deltaup{+8.4}$ &
$\textbf{15.3}\pm 4.8\,\deltaup{+8.3}$ &
$\textbf{87.5}\pm 8.3\,\deltaup{+15.3}$ &
$\textbf{79.2}\pm 4.8\,\deltaup{+29.2}$ &
$\textbf{72.2}\pm 2.4\,\deltaup{+15.2}$ &
$\textbf{41.0}\pm 2.4\,\deltaup{+17.7}$ \\

\midrule
Cosmos-Predict2~\cite{agarwal2025cosmos}-\textit{Few}~+~\fa{}~\&~\rl{} &
$\textbf{69.5}\pm 8.7\,\deltaup{+23.6}$ &
$\textbf{26.4}\pm 6.4\,\deltaup{+8.4}$ &
$48.6\pm 6.4\,\deltaup{+8.3}$ &
$\textbf{38.9}\pm 2.4\,\deltaup{+20.8}$ &
$61.1\pm 4.8\,\deltadown{-8.3}$ &
$9.6\pm 2.5\,\deltaup{+2.6}$ &
$76.4\pm 8.6\,\deltaup{+4.4}$ &
$54.2\pm 8.4\,\deltaup{+4.2}$ &
$63.9\pm 3.2\,\deltaup{+6.9}$ &
$32.3\pm 2.2\,\deltaup{+9.0}$ \\

\wm-\textit{Few}~+~\fa{}~\&~\rl{} &
$55.6\pm 6.4\,\deltaup{+9.7}$ &
$\textbf{26.4}\pm 2.4\,\deltaup{+8.4}$ &
$\textbf{51.4}\pm 2.4\,\deltaup{+11.1}$ &
$33.3\pm 4.2\,\deltaup{+15.2}$ &
$\textbf{70.8}\pm 4.2\,\deltaup{+1.4}$ &
$\textbf{11.1}\pm 4.8\,\deltaup{+4.1}$ &
$\textbf{88.9}\pm 4.8\,\deltaup{+16.7}$ &
$\textbf{75.0}\pm 8.3\,\deltaup{+25.0}$ &
$\textbf{66.0}\pm 3.1\,\deltaup{+9.0}$ &
$\textbf{36.5}\pm 1.8\,\deltaup{+13.2}$ \\

\bottomrule
\end{tabular}
}
\label{tab:simplerenv_windowx_wm_ablation}
\end{table}

%% file: tables/simplerenv_windowx_num_demo_ablation.tex
\begin{table}[!htbp]
\centering
\caption{\textbf{Few-shot RL in world model.}
Ablation on the four SimplerEnv-WidowX tasks with VLA-Adapter-0.5B (SFT 200k steps), where post-training with \fa{} and \rl{} in \wm{} is initialized from only a small number of seed trajectories per task (100-\textit{img.} \textit{vs.} 10-\textit{img.}).
}
\resizebox{1\textwidth}{!}{
\begin{tabular}{l|cccccccccc}
\toprule
\multirow{2}{*}{Method} &
\multicolumn{2}{c}{\textbf{Put Spoon on Towel}} &
\multicolumn{2}{c}{\textbf{Put Carrot on Plate}} &
\multicolumn{2}{c}{\textbf{Stack Green Block on Yellow Block}} &
\multicolumn{2}{c}{\textbf{Put Eggplant in Yellow Basket}} &
\textbf{Partial} &
\textbf{Overall} \\
& Grasp Spoon & Success & Grasp Carrot & Success & Grasp Green Block & Success & Grasp Eggplant & Success & Average & Average \\
\midrule
VLA-Adapter-0.5B~\cite{wang2025vla} &
$45.9\pm 7.2$ & $18.0\pm 8.4$ & $40.3\pm 6.3$ & $18.1\pm 4.8$ &
$69.4\pm 6.3$ & $7.0\pm 4.8$ & $72.2\pm 7.2$ & $50.0\pm 4.8$ &
$57.0\pm 5.1$ & $23.3\pm 2.2$ \\
\hspace{0.9em}+\,\fa{}~\&~\rl{}\textit{--10-img.} &
$51.4\pm 2.4\,\deltaup{+5.5}$ &
$26.4\pm 2.4\,\deltaup{+8.4}$ &
$45.9\pm 7.2\,\deltaup{+5.6}$ &
$27.8\pm 6.4\,\deltaup{+9.7}$ &
$72.2\pm 10.5\,\deltaup{+2.8}$ &
$\textbf{18.0}\pm 4.8\,\deltaup{+11.0}$ &
$84.7\pm 2.4\,\deltaup{+12.5}$ &
$66.7\pm 4.2\,\deltaup{+16.7}$ &
$66.7\pm 4.2\,\deltaup{+9.7}$ &
$34.7\pm 3.9\,\deltaup{+11.4}$ \\

\hspace{0.9em}+\,\fa{}~\&~\rl{}\textit{--100-img.} &
$\textbf{70.8}\pm 7.2\,\deltaup{+24.9}$ &
$\textbf{33.3}\pm 4.2\,\deltaup{+15.3}$ &
$\textbf{52.8}\pm 2.4\,\deltaup{+12.5}$ &
$\textbf{36.1}\pm 4.8\,\deltaup{+18.0}$ &
$\textbf{77.8}\pm 4.8\,\deltaup{+8.4}$ &
$15.3\pm 4.8\,\deltaup{+8.3}$ &
$\textbf{87.5}\pm 8.3\,\deltaup{+15.3}$ &
$\textbf{79.2}\pm 4.8\,\deltaup{+29.2}$ &
$\textbf{72.2}\pm 2.4\,\deltaup{+15.2}$ &
$\textbf{41.0}\pm 2.4\,\deltaup{+17.7}$ \\
\bottomrule
\end{tabular}
}
\label{tab:simplerenv_windowx_num_demo_ablation}
\end{table}

%% file: tables/real_world_exp.tex
\begin{table}[!htbp]
\centering
\caption{\textbf{Real-robot world model with RL evaluation on UR30e.}
All VLA policies are first SFT 50k steps, then post-trained for 100 RL steps with \fa{} and \rl{} in \wm{}, and evaluated on the real robot.}
\resizebox{0.85\columnwidth}{!}{
\begin{tabular}{l|ccccc}
\toprule
\textbf{Method} & \textbf{GraspBottle} & \textbf{PlaceBowl} & \textbf{PulloutTissue} & \textbf{PlaceCube} & \textbf{Overall} \\
\midrule
VLA-Adapter-0.5B~\cite{wang2025vla} &
$45.0\pm 8.7$ &
$13.3\pm 5.8$ &
$28.3\pm 14.4$ &
$56.7\pm 5.8$ &
$35.8\pm 3.1$ \\
\hspace{0.9em}+\,\fa~\&~\rl &
$\textbf{76.7}\pm 2.9\,\deltaup{+31.7}$ &
$\textbf{46.7}\pm 7.6\,\deltaup{+33.4}$ &
$\textbf{51.7}\pm 5.8\,\deltaup{+23.4}$ &
$\textbf{66.7}\pm 5.8\,\deltaup{+10.0}$ &
$\textbf{60.4}\pm 0.7\,\deltaup{+24.6}$ \\
\midrule
Pi0.5-3B~\cite{black2025pi_} &
$58.3\pm 2.9$ &
$51.7\pm 5.8$ &
$33.3\pm 5.8$ &
$65.0\pm 5.0$ &
$52.1\pm 3.8$ \\
\hspace{0.9em}+\,\fa~\&~\rl &
$\textbf{86.7}\pm 2.9\,\deltaup{+28.4}$ &
$\textbf{83.3}\pm 2.9\,\deltaup{+31.6}$ &
$\textbf{66.7}\pm 2.9\,\deltaup{+33.4}$ &
$\textbf{91.7}\pm 2.9\,\deltaup{+26.7}$ &
$\textbf{82.1}\pm 0.7\,\deltaup{+30.0}$ \\
\midrule
OpenVLA-OFT-7B~\cite{kim2025openvla} &
$55.0\pm 5.0$ &
$33.3\pm 2.9$ &
$41.7\pm 2.9$ &
$11.7\pm 2.9$ &
$35.4\pm 0.7$ \\
\hspace{0.9em}+\,\fa~\&~\rl &
$\textbf{73.3}\pm 2.9\,\deltaup{+18.3}$ &
$\textbf{50.0}\pm 5.0\,\deltaup{+16.7}$ &
$\textbf{81.7}\pm 2.9\,\deltaup{+40.0}$ &
$\textbf{46.7}\pm 2.9\,\deltaup{+35.0}$ &
$\textbf{62.9}\pm 0.7\,\deltaup{+27.5}$ \\
\bottomrule
\end{tabular}
}
\label{tab:real_world_exp}
\end{table}

%% file: images/new_behaviour_demo.tex
\begin{figure}[H]
  \centering
  \includegraphics[width=0.8\textwidth]{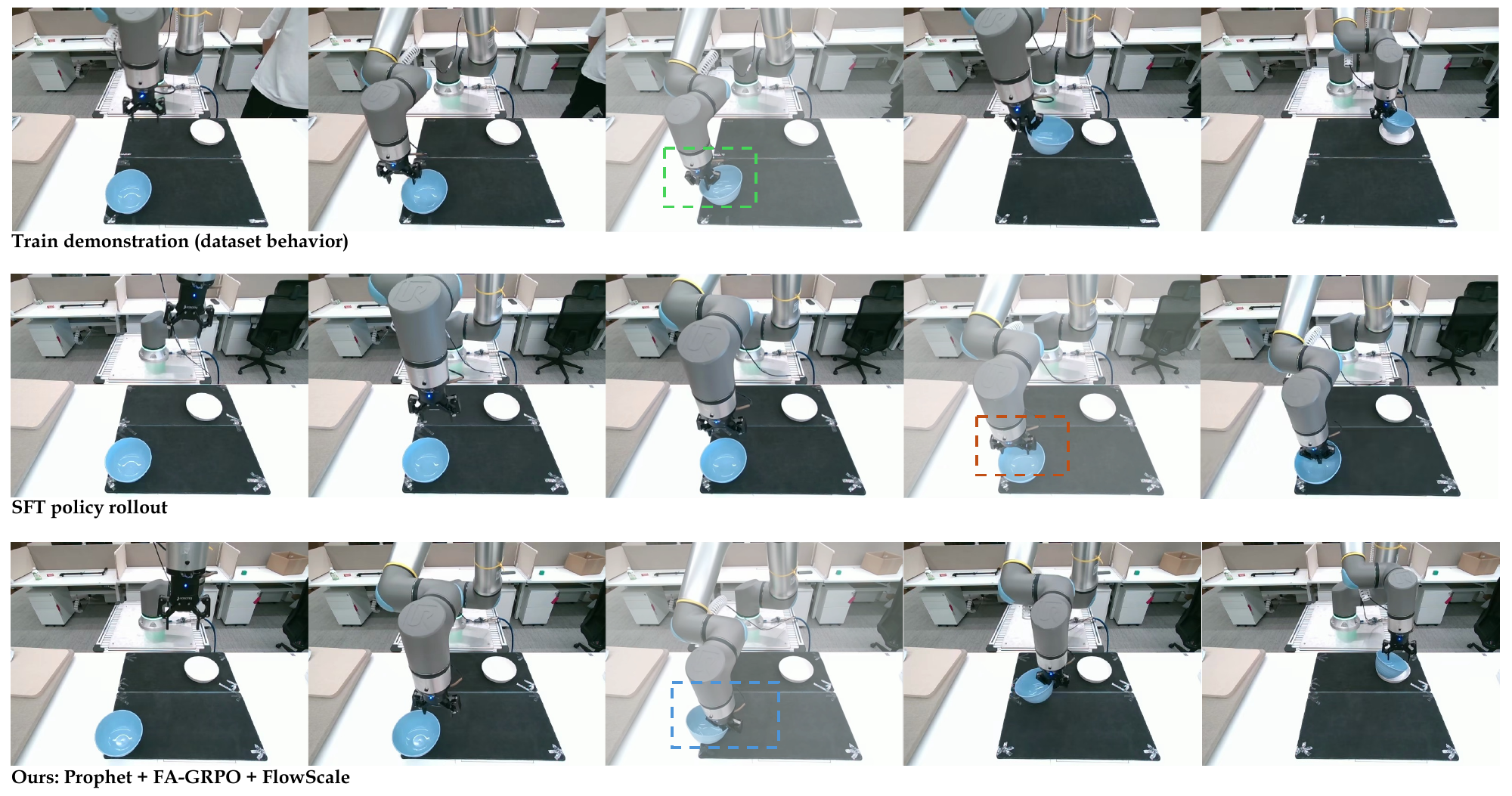}
\caption{\textbf{Real-world rollouts on the \emph{PlaceBowl} task.} \textbf{Top:} examples from collected training data, where the gripper approaches and grasps the bowl from the left side. 
\textbf{Middle:} failed manipulation by the SFT policy (Pi0.5), which inherits the left-side approach but cannot complete the task. 
\textbf{Bottom:} policy after post-training with \fa{} and \rl{} in \wm{}, which learns a new and consistent right-side approach not 
present in the demonstrations.
}
  \label{fig:rl_demo_pb}
\end{figure}

%% file: images/rl_tissue.tex
\begin{figure}[!htbp]
  \centering
  \includegraphics[width=0.8\linewidth]{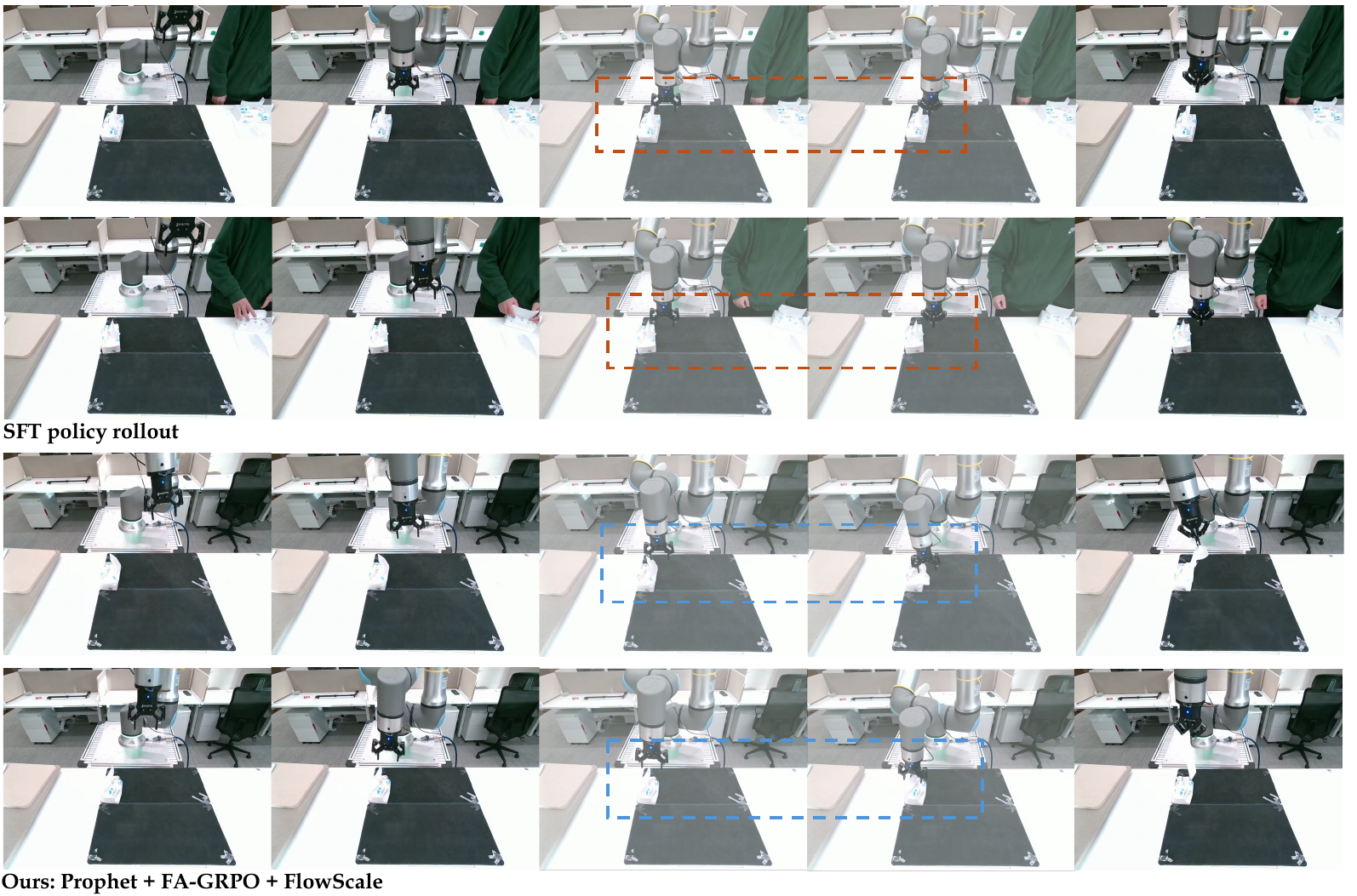}
\caption{\textbf{Real-world rollouts on the \emph{PulloutTissue} task.}
    The first two rows show rollouts from the SFT policy (Pi0.5), which often drifts laterally when approaching the exposed tissue edge.
    As highlighted, the gripper frequently deviates from the intended pulling direction, leading to missed grasps or weak contact with the tissue.
    The bottom two rows show our policy after post-training with \fa{} and \rl{} in \wm{}, which produces a much more stable approach and consistently aligns the gripper with the tissue edge, resulting in reliable extraction and placement.
    This illustrates that RL can correct soft-object manipulation behaviors that remain brittle under SFT alone.
}
  \label{fig:rl_demo_pt}
\end{figure}

%% file: images/rl_wm_rollout_tissue.tex
\begin{figure}[!htbp]
  \centering
  \includegraphics[width=0.8\linewidth]{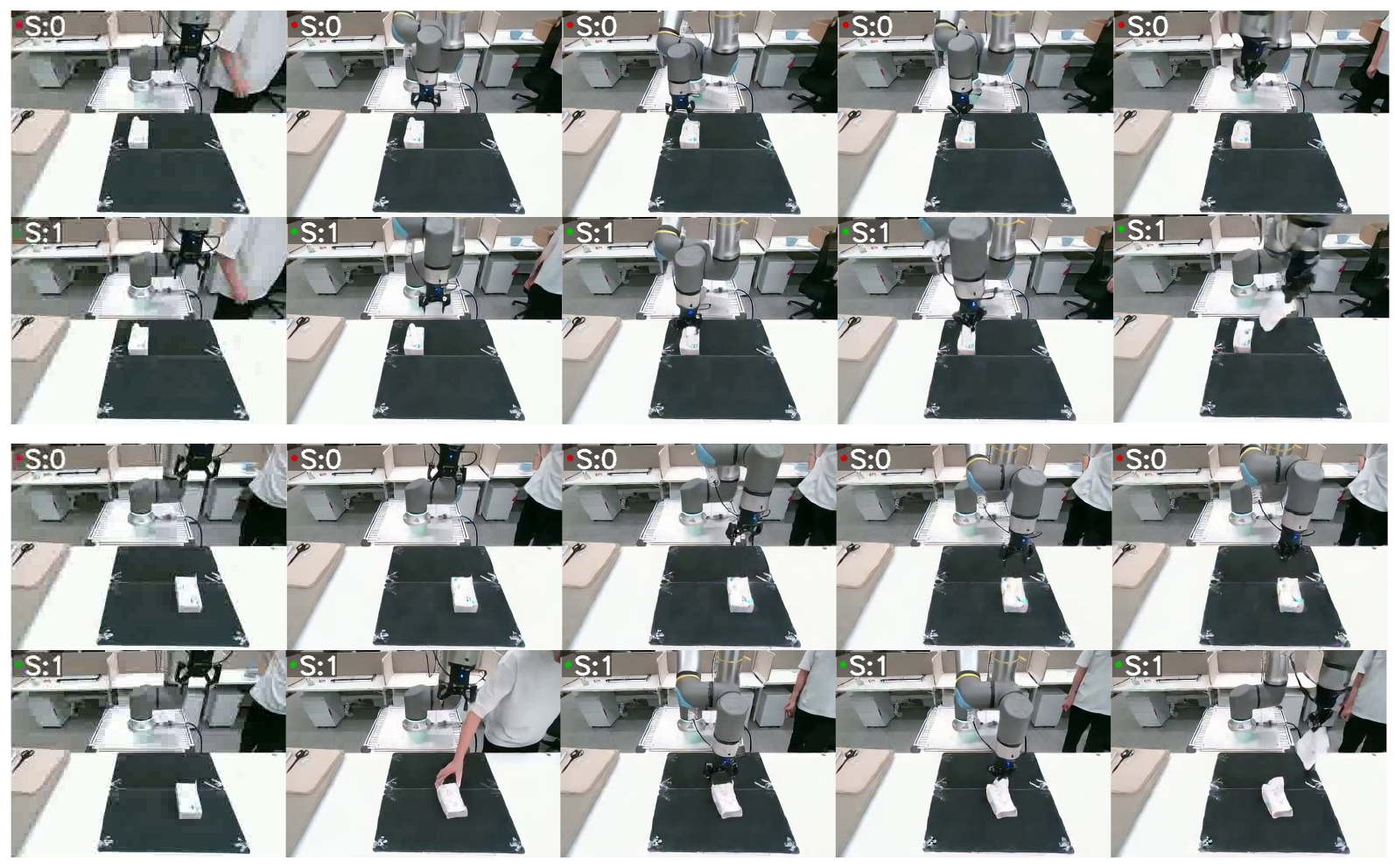}
    \caption{\textbf{\wm{} rollouts during RL training on \emph{PulloutTissue}.}
We show successful (S:1) and failed (S:0) \wm{} rollouts from two initial object positions. Each sequence is the full predicted manipulation trajectory used for policy optimization, with the top-left label giving the RM’s majority-vote decision. These examples illustrate the variability of the \wm{} and how both successes and failures influence the policy during RL.
    }
  \label{fig:rl_demo_wm_rollout_tissue}
\end{figure}

%% file: tables/rl_ablation_libero.tex
\begin{table}[!htbp]
\centering
\setlength{\tabcolsep}{6pt}
\caption{\textbf{RL in simulator \textit{vs.} RL in world model (\wm{}) on LIBERO.}
Green numbers denote absolute gains over VLA-Adapter (SFT 10k steps).
Simulator rows use subscripts to mark the first RL update where the peak validation score is reached. \wm{} rows omit subscripts since we select the best checkpoint under a different budget.}

\small
\resizebox{\linewidth}{!}{
\begin{tabular}{l c c c c c}
\toprule
\textbf{Method}
& \textbf{Spatial}
& \textbf{Object}
& \textbf{Goal}
& \textbf{Long}
& \textbf{Overall}
\\
\midrule
VLA-Adapter~\cite{wang2025vla}
  & $82.7 \pm 0.9$ &
    $78.3 \pm 1.3$ &
    $80.0 \pm 0.2$ &
    $78.6 \pm 2.0$ &
    $79.9 \pm 2.2$
\\
\midrule
\multicolumn{6}{l}{\textit{Simulator only}} \\
\midrule
\hspace{0.9em}+\,\fa
  & $92.4 \pm 0.5\,\deltastep{+9.7}{409}$ &
    $86.9 \pm 0.7\,\deltastep{+8.6}{269}$ &
    $87.4 \pm 1.4\,\deltastep{+7.4}{259}$ &
    $84.4 \pm 1.2\,\deltastep{+5.8}{389}$ &
    $87.8 \pm 3.2\,\deltaup{+7.9}$
\\
\hspace{0.9em}+\,\fa~\&~\rl
  & $\textbf{94.6} \pm 1.2\,\deltastep{+11.9}{119}$ &
    $\textbf{87.4} \pm 0.5\,\deltastep{+9.1}{159}$ &
    $\textbf{91.2} \pm 0.7\,\deltastep{+11.2}{179}$ &
    $\textbf{86.4} \pm 0.6\,\deltastep{+7.8}{169}$ &
    $\textbf{90.1} \pm 3.5\,\deltaup{+10.2}$
\\
\midrule
\multicolumn{6}{l}{\textit{Model only (\wm{})}} \\
\midrule
\hspace{0.9em}+\,\fa
  & $85.2 \pm 0.4\,\deltaup{+2.5}$ &
    $80.6 \pm 1.1\,\deltaup{+2.3}$ &
    $82.9 \pm 0.5\,\deltaup{+2.9}$ &
    $80.6 \pm 1.0\,\deltaup{+2.0}$ &
    $82.3 \pm 0.7\,\deltaup{+2.9}$
\\
\hspace{0.9em}+\,\fa~\&~\rl
  & $\textbf{89.0} \pm 1.1\,\deltaup{+6.3}$ &
    $\textbf{81.9} \pm 1.3\,\deltaup{+3.6}$ &
    $\textbf{83.7} \pm 1.1\,\deltaup{+3.7}$ &
    $\textbf{83.6} \pm 1.3\,\deltaup{+3.0}$ &
    $\textbf{84.5} \pm 1.1\,\deltaup{+5.1}$
\\
\bottomrule
\end{tabular}
}
\label{tab:rl_ablation}
\end{table}

%% file: tables/flowscale_sensitivity.tex
\begin{table}[!htbp]
\centering
\begin{minipage}[c]{0.55\linewidth}
\centering
\scriptsize
\setlength{\tabcolsep}{3pt}
\renewcommand{\arraystretch}{0.92}
\caption{\textbf{\rl{} sensitivity on LIBERO Spatial.}
We report success gain over the SFT policy and the first RL update that reaches the peak validation score.}
\begin{tabular}{lcccc}
\toprule
\textbf{Setting} & $p$ & $\alpha$ & Clip & SuccGain$\uparrow$ / Peak$\downarrow$ \\
\midrule
w/o \rl{} & 0.0 & --  & --        & 9.7 / 409 \\
Default      & 0.5 & 0.3 & 0.3--2.0  & 11.9 / 119 \\
$p$ low      & 0.3 & 0.3 & 0.3--2.0  & 10.7 / 170 \\
$p$ high     & 0.7 & 0.3 & 0.3--2.0  & 12.9 / 109 \\
$\alpha$ low & 0.5 & 0.1 & 0.3--2.0  & 11.1 / 139 \\
$\alpha$ high& 0.5 & 0.5 & 0.3--2.0  & 10.9 / 129 \\
Tight clip   & 0.5 & 0.3 & 0.6--1.6  & 10.1 / 49  \\
\bottomrule
\end{tabular}
\label{tab:flowscale_sens}
\end{minipage}
\hfill
\begin{minipage}[c]{0.40\linewidth}
\centering
\scriptsize
\setlength{\tabcolsep}{4pt}
\renewcommand{\arraystretch}{0.92}
\caption{\textbf{Reward control on LIBERO Spatial simulator RL.}
All rows use VLA-Adapter in the simulator-only setting of Tab.~\ref{tab:rl_ablation}.}
\begin{tabular}{lc}
\toprule
\textbf{Setting} & Success$\uparrow$ \\
\midrule
VLA-Adapter SFT & $82.7 \pm 0.9$ \\
Simulator RL w/ GT reward & $94.6 \pm 1.2$ \\
Simulator RL w/ 72B RM & $90.4 \pm 0.6$ \\
\bottomrule
\end{tabular}
\label{tab:sim_rm_control}
\end{minipage}
\end{table}

%% file: tables/rm_ablation.tex
\begin{table}[!htbp]
\centering
\caption{\textbf{RM ablation on \textsc{PutCarrot} task.}
Success gain, latency, and GPU hours under vLLM inference with batch size 256 are reported.}
\label{tab:rm_cost}
\begin{minipage}[t]{0.48\linewidth}
\centering
\scriptsize
\setlength{\tabcolsep}{4pt}
\renewcommand{\arraystretch}{0.95}
\begin{tabular}{cccc}
\toprule
\textbf{RM} & SuccGain$\uparrow$ & Lat.$\downarrow$ & GPUh$\downarrow$ \\
\midrule
7B  &  5.8 & 0.16 &  7.3 \\
32B &  3.4 & 0.66 & 30.2 \\
72B & 13.8 & 0.74 & 34.1 \\
\bottomrule
\end{tabular}

{\footnotesize \textbf{(a)} Size sweep (Qwen2.5-VL).}
\end{minipage}
\hfill
\begin{minipage}[t]{0.48\linewidth}
\centering
\scriptsize
\setlength{\tabcolsep}{3pt}
\renewcommand{\arraystretch}{0.95}
\begin{tabular}{ccccc}
\toprule
\textbf{RM} & Vote & SuccGain$\uparrow$ & Lat.$\downarrow$ & GPUh$\downarrow$ \\
\midrule
72B & 1 &  6.9 & 0.33 & 15.1 \\
72B & 3 &  9.7 & 0.42 & 24.3 \\
72B & 5 & 13.8 & 0.74 & 34.1 \\
\bottomrule
\end{tabular}

{\footnotesize \textbf{(b)} Voting sweep (72B).}
\end{minipage}
\end{table}

%% file: images/reward_curve.tex
\begin{figure*}[t]
  \centering
  \includegraphics[width=\textwidth]{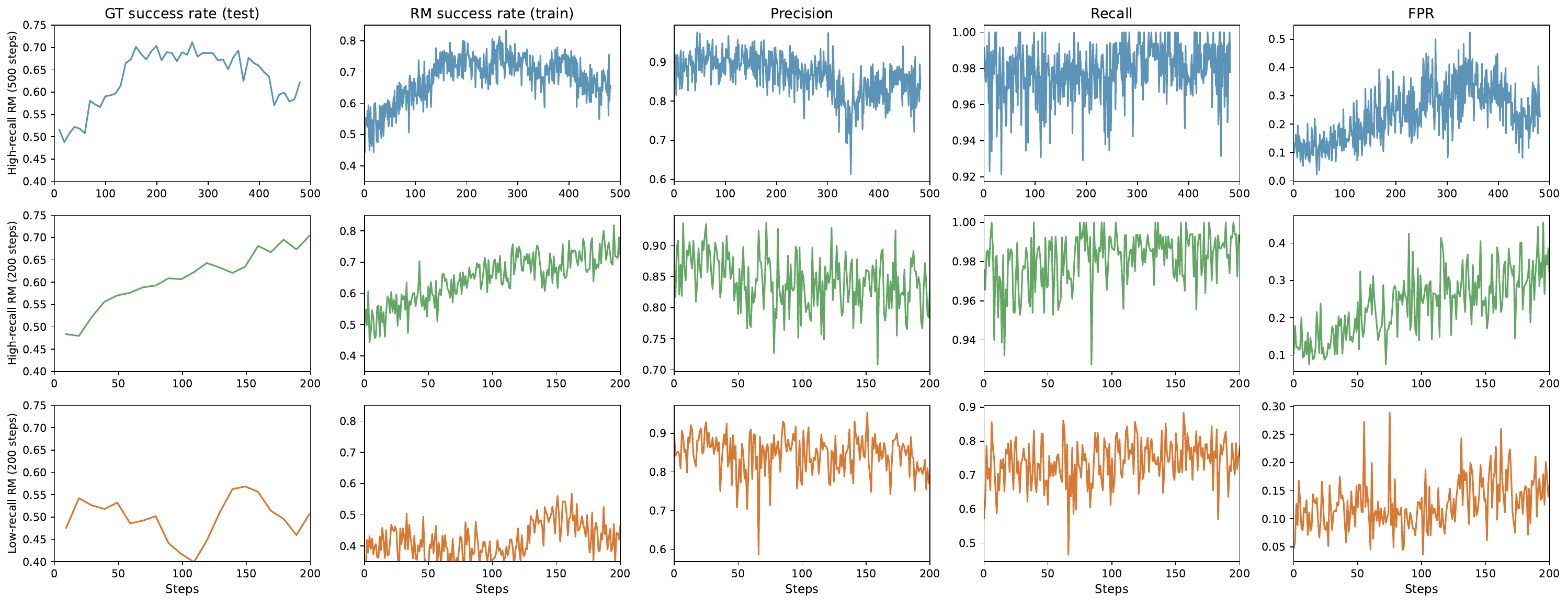}
  \caption{
  \textbf{RM diagnostics for three SmolVLA with \fa{} runs on LIBERO-Spatial.}
    Columns show (left to right) test success rate under the true simulator reward,
    average RM score on on-policy rollouts, RM precision, RM recall (TPR), and RM false positive rate (FPR).
    \textbf{Top:} high-recall RM, long run (500 steps), where success first improves and then collapses as precision drops sharply.
    \textbf{Middle:} high-recall RM, short run (200 steps), where precision/recall stay high and success improves monotonically.
    \textbf{Bottom:} low-recall RM, short run (200 steps), where FPR is low and precision is comparable but recall is much lower, so the policy fails to improve.
  }
  \label{fig:rm_diagnostics}
\end{figure*}

%% file: sections/5-conclusion.tex
\section{Conclusions}
In this paper, we studied how to make VLA post-training effective and practical by coupling policies with an adaptive world model. 
We introduced \wm{}, an action-conditioned video world model that generates long-horizon, action-aligned manipulation rollouts from first-frame observations and multi-step actions, and showed that large-scale pretraining with few-shot adaptation yields a simulator that transfers across robots, objects, and environments. 
Building on \wm{}, we proposed \rl{}, a flow-aware GRPO variant with stabilized gradients for reliable long-horizon RL with the world model loop.
Across diverse VLA variants, our experiments show success gains of 5--17\% on public benchmarks and 24--30\% on real-robot evaluations.

At the same time, the current system is computationally demanding: during RL, the policy must interact with a 2B-parameter \wm{} to generate closed-loop rollouts, which dominates training cost and limits feasible iterations. Improving world-model efficiency—for example through architectural simplification, student distillation, cross-rollout feature caching, or specialized inference kernels—could substantially accelerate RL and enable scaling to longer horizons, larger task suites, and richer policy exploration.